\documentclass{article}

\PassOptionsToPackage{super,sort&compress,comma,compress}{natbib}

\usepackage[preprint]{neurips_2021}

\usepackage[utf8]{inputenc} % allow utf-8 input
\usepackage[T1]{fontenc}    % use 8-bit T1 fonts
\usepackage[hidelinks]{hyperref}       % hyperlinks
\usepackage{url}            % simple URL typesetting
\usepackage{booktabs}       % professional-quality tables
\usepackage{amsfonts}       % blackboard math symbols
\usepackage{nicefrac}       % compact symbols for 1/2, etc.
\usepackage{microtype}      % microtypography
\usepackage{xcolor}         % colors
\usepackage{graphicx}
\usepackage{caption}
\usepackage{subcaption}
\usepackage{multirow}
\usepackage{multicol}
\usepackage[normalem]{ulem} % for sout (strikethrough)
\usepackage{wrapfig}
\usepackage[percent]{overpic}
\usepackage{lipsum}
\usepackage{csquotes}
\usepackage[OT2,T1]{fontenc}
\usepackage[russian,greek,english]{babel}
\DeclareUnicodeCharacter{FFFD}{?}

\newcommand{\eqn}[1]{\begin{equation}#1\end{equation}}

\newif\ifcomment
\title{Video PreTraining (VPT): Learning to Act by Watching Unlabeled Online Videos}

\author{
Bowen Baker\footnotemark[1]\hspace{4px}\footnotemark[2]\\bowen@openai.com\And
Ilge Akkaya\footnotemark[1]\hspace{4px}\footnotemark[2]\\ilge@openai.com\And
Peter Zhokhov\footnotemark[1]\hspace{4px}\footnotemark[2]\\peterz@openai.com\And
Joost Huizinga\footnotemark[1]\hspace{4px}\footnotemark[2]\\joost@openai.com\And
Jie Tang\footnotemark[1]\hspace{4px}\footnotemark[2]\\jietang@openai.com\And
Adrien Ecoffet\footnotemark[1]\hspace{4px}\footnotemark[2]\\adrien@openai.com\And
Brandon Houghton\footnotemark[1]\hspace{4px}\footnotemark[2]\\brandon@openai.com\And
Raul Sampedro\footnotemark[1]\hspace{4px}\footnotemark[2]\\raulsamg@gmail.com\And
Jeff Clune\thanks{This was a large effort by a dedicated team. Each author made huge contributions on many fronts over long time periods. All members were full time on the project for over six months. BB, IA, PZ, and JC were on the original VPT project team and were thus involved for even longer (over a year). Aside from those original team members, author order is random. It was also randomized between IA and PZ.}\hspace{4px}\thanks{OpenAI}\hspace{4px}\thanks{University of British Columbia}\\jclune@gmail.com
}

\begin{document}
\commenttrue % Set \commentfalse to togle inline coments off

\maketitle

\renewcommand{\thefootnote}{(\roman{footnote})}

\begin{abstract}
Pretraining on noisy, internet-scale datasets has been heavily studied as a technique for training models with broad, general capabilities for text, images, and other modalities.\cite{brown2020language,devlin2018bert,liu2019roberta,raffel2019exploring,mahajan2018exploring,radford2021learning}
However, for many sequential decision domains such as robotics, video games, and computer use, publicly available data does not contain the labels required to train behavioral priors in the same way. We extend the internet-scale pretraining paradigm to sequential decision domains through semi-supervised imitation learning wherein agents learn to act by watching online unlabeled videos.
Specifically, we show that with a small amount of labeled data we can train an inverse dynamics model accurate enough to label a huge unlabeled source of online data -- here, online videos of people playing Minecraft -- from which we can then train a general behavioral prior.
Despite using the native human interface (mouse and keyboard at 20Hz), we show that this behavioral prior has nontrivial zero-shot capabilities and that it can be fine-tuned, with both imitation learning and reinforcement learning, to hard-exploration tasks that are impossible to learn from scratch via reinforcement learning. For many tasks our models exhibit human-level performance, and we are the first to report computer agents that can craft diamond tools, which can take proficient humans upwards of 20 minutes (24,000 environment actions) of gameplay to accomplish. 
\end{abstract}

\section{Introduction}

Work in recent years has demonstrated the efficacy of pretraining large and general foundation models \cite{bommasani2021opportunities} on noisy internet-scale datasets for use in downstream tasks in natural language \cite{devlin2018bert,liu2019roberta,raffel2019exploring,brown2020language} and computer vision. \cite{mahajan2018exploring, zhai2021scaling, radford2021learning}
For sequential decision domains (e.g.\ robotics, game playing, and computer usage) where agents must repeatedly act within an environment, a wealth of data also exists on the web; however, most of this data is in the form of \textit{unlabeled} video (i.e. without the actions taken at each frame), making it much less straightforward to train a behavioral prior in these domains than it is in e.g.\ natural language.
In a few rare settings, such as Chess, Go, and StarCraft, there already exist large datasets with action labels from various online platforms that researchers have used for imitation learning. \cite{silver2016mastering, vinyals2019grandmaster}
When large labeled datasets do not exist, the canonical strategy for training capable agents is reinforcement learning (RL),\cite{sutton2018reinforcement} which can be sample inefficient and expensive for hard-exploration problems.\cite{berner2019dota, baker2019emergent, jaderberg2019human, badia2020agent57, bellemare2016unifying, burda2018exploration, ecoffet2021first}
Many virtual tasks, e.g.\ navigating websites, using Photoshop, booking flights, etc., can be very hard to learn with RL and do not have large, commonly available sources of labeled data.\cite{humphreys2022data,pmlrv70shi17a}
In this paper, we seek to extend the paradigm of training large, general-purpose foundation models to sequential decision domains by utilizing freely available internet-scale unlabeled video datasets with a simple semi-supervised imitation learning method. We call this method Video PreTraining (VPT) and demonstrate its efficacy in the domain of Minecraft.

Existing semi-supervised imitation learning methods aim to learn with few or no explicit action labels; however, they generally rely on the policy's ability to explore the environment throughout training, making them susceptible to exploration bottlenecks. \cite{ng2000algorithms, torabi2019recent, ho2016generative, torabi2018behavioral, liu2018imitation}
Furthermore, most prior semi-supervised imitation learning work was tested in the relatively low data regime; because we experiment with \textit{far} more data ($\sim$70k hours of unlabeled video), we hypothesize that we can achieve good performance with a much simpler method, a trend that has proven true for pretraining in other modalities such as  text.\cite{brown2020language}
In particular, given a large but unlabeled dataset, we propose generating pseudo-labels by gathering a small amount of labeled data to train an inverse dynamics model (IDM) that predicts the action taken at each timestep in a video.
Behavioral cloning (BC) can require a large amount of data because the model must learn to infer intent and the distribution over future behaviors from only past observations.
In contrast, the inverse dynamics modeling task is simpler because it is \emph{non-causal}, meaning it can look at both past and future frames to infer actions.
In most settings, environment mechanics are far simpler than the breadth of human behavior that can take place within the environment, suggesting that non-causal IDMs could require far less data to train than causal BC models.
Using pseudo-labels generated from the IDM, we then train a model to mimic the distribution of behavior in the previously unlabeled dataset with standard behavioral cloning at scale, which does not require any model rollouts and thus does not suffer from any potential exploration bottlenecks in the environment.
Finally, we show we can fine-tune this model to downstream tasks with either behavioral cloning or reinforcement learning.

\begin{wrapfigure}{r}{0.33\textwidth}
\vspace{-20px}
\begin{center}
\includegraphics[width=\linewidth]{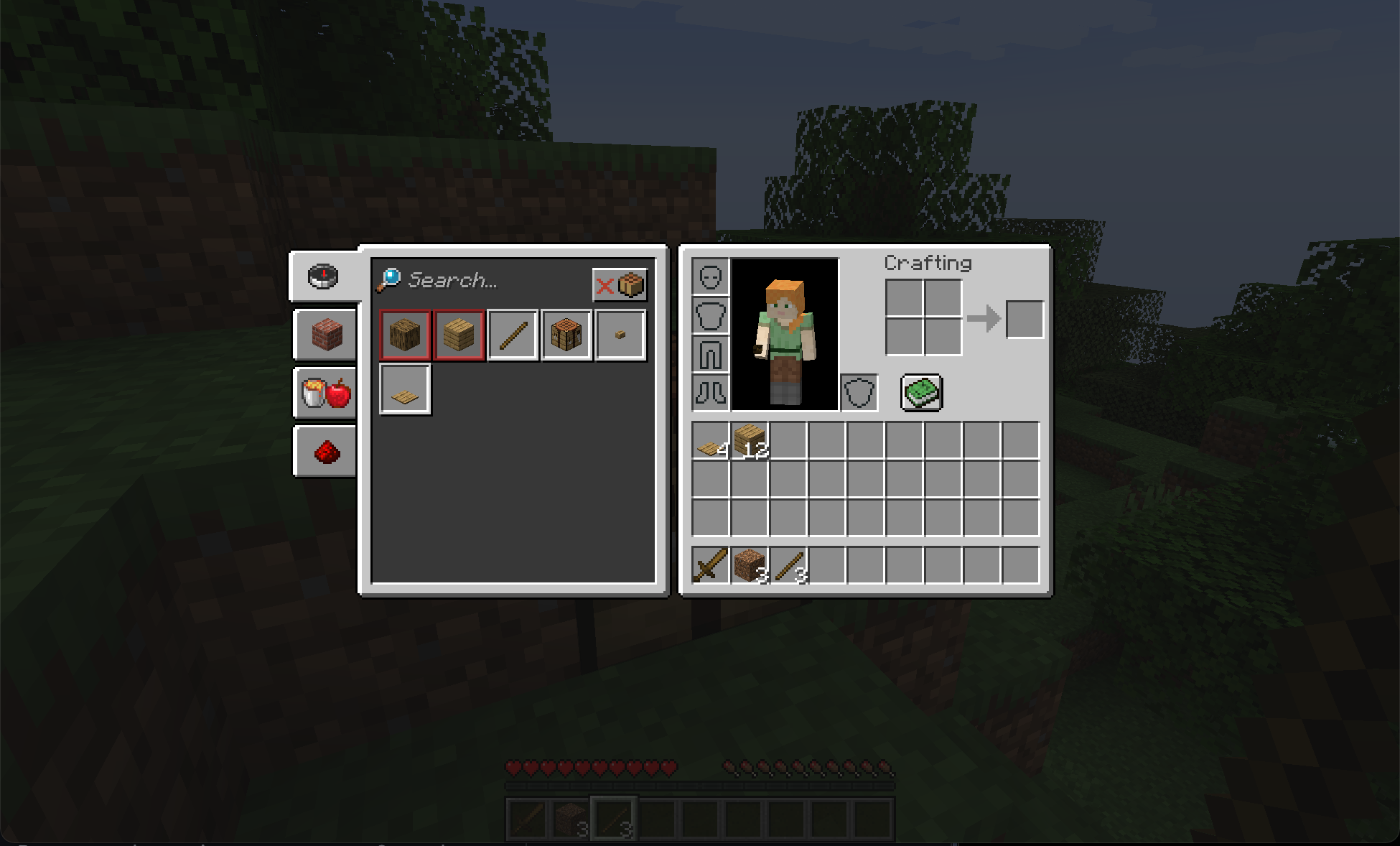} 
\end{center}
\vspace{-8px}
\caption{Example Minecraft crafting GUI. Agents use the mouse and keyboard to navigate menus and drag and drop items.
}
\label{fig:crafting_window}
\vspace{-10px}
\end{wrapfigure}
We chose to test our method in Minecraft because (a) it is one of the most actively played games in the world\cite{twinfinite2021played} and thus has a wealth of commonly available video data online, (b) it is a fairly open-ended sandbox game with an extremely wide variety of potential things to do, build, and collect, making our results more applicable to real-world applications such as computer usage, which also tends to be varied and open-ended, and (c) it has already garnered interest by the RL community as a research domain due to its complexity and correspondingly difficult exploration challenges.\cite{guss2019minerl, tessler2017deep, scheller2020sample, kanitscheider2021multi, oh2016control}
In this work we use the native human interface for Minecraft so that we can (1) most accurately model the human behavior distribution and reduce domain shift between video data and the environment, (2) make data collection easier by allowing our human contractors to play the game without modification, and (3) eliminate the need to hand-engineer a custom interface for models to interact with the environment.
This choice means that our models play at 20 frames per second and must use a mouse and keyboard interface to interact with human GUIs for crafting, smelting, trading, etc., including dragging items to specific slots or navigating the recipe book with the mouse cursor (Fig.~\ref{fig:crafting_window}).
Compared to prior work in Minecraft that uses a lower frame rate and constructs crafting and attacking macros,\cite{kanitscheider2021multi, patil2020align, skrynnik2020forgetful, lin2021juewu} using the native human interface drastically increases the environment's exploration difficulty, making most simple tasks near impossible with RL from scratch.
% which we validate in Sec.~\ref{section:results}.
Even the simple task of gathering a single wooden log while already facing a tree takes 60 consecutive attack actions with the human interface, meaning the chance for a naive random policy to succeed is $\frac{1}{2}^{60}$. While this paper shows results in Minecraft only, the VPT method is general and could be applied to any domain.

In Section \ref{section:results} we show that the VPT foundation model has nontrivial zero-shot performance, accomplishing tasks impossible to learn with RL alone, such as crafting planks and crafting tables (tasks requiring a human proficient in Minecraft a median of 50 seconds or $\sim$970 consecutive actions).
Through fine-tuning with behavioral cloning to smaller datasets that target more specific behavior distributions, our agent is able to push even further into the technology tree, crafting stone tools (taking a human a median of 2.3 minutes or $\sim$2790 actions).
Finally, fine-tuning via RL produces the most dramatic improvements: our agent is able to craft diamond tools, an unprecedented result in Minecraft made even more challenging by using the native human interface. This task requires a proficient human a median upwards of 20 minutes or $\sim$24000 actions.
The main contributions of this work are (1) we are the first to show promising results applying semi-supervised imitation learning to extremely large, noisy, and freely available video datasets for sequential decision domains, (2) we show that such pretraining plus fine-tuning enables agents to solve tasks that were otherwise impossible to learn, (3) we show that labeled contractor data is far more efficiently used within the VPT method than it would be by directly training a foundation model from it and (4) we open source our contractor data, trained model weights, and Minecraft environment for future research into learning to act via semi-supervised imitation learning at scale.
\section{Preliminaries and Related Work}
\label{section:related}
\vspace{-2px}

Imitation learning methods \cite{pomerleau1988alvinn, schaal1999imitation, argall2009survey, hussein2017imitation} seek to construct a policy that accurately models the distribution of behavior in some dataset $D = \{(o_i, a_i)\}, ~ i \in \{1...N\}$ of action-observation pairs. In order to roll out these policies in an environment, they must be \emph{causal}, meaning they condition on observations from the current timestep $t$ and past timesteps only, i.e. $\pi \sim p(a_t | o_1...o_t)$. Imitation learning is simplest when demonstrations are labeled with corresponding actions. Imitating labeled trajectories has seen success in aerial vehicles, \cite{sammut1992learning, giusti2015machine} self-driving cars, \cite{bojarski2016end, codevilla2018end} board games, \cite{coulom2007computing, silver2016mastering} and video games.\cite{hester2018deep, vinyals2019grandmaster}

When labeled demonstrations are not available, standard behavioral cloning will not work; however, there is a large body of work in imitating behavior from unlabeled demonstrations. \cite{torabi2019recent} For instance, GAIL \cite{ho2016generative} constructs an adversarial objective incentivizing the trained policy to exhibit behaviors indistinguishable from those in the target dataset.
\citet{edwards2019imitating} propose to first learn a latent policy using unlabeled demonstrations and then map the learned latent actions to real actions with a small amount of environment interaction.
\citet{peng2018sfv} first use motion-capture methods to track agent positions in videos and then train RL agents to match these waypoints.
Similarly, \citet{behbahani2019learning} and \citet{aytar2018playing} task a RL agent to match waypoints; however, they construct waypoints that are embeddings from unsupervised feature learning models. 
\citet{pathakICLR18zeroshot} and \citet{nairICRA2017combine} train goal conditioned policies to take actions that advance the current state towards expert-provided goal states expressed as high dimensional visual waypoints.
Most similar to our own work, \citet{torabi2018behavioral} simultaneously train (1) an inverse dynamics model (IDM),\cite{nguyen2008learning} which aims to uncover the underlying action between timesteps given observations of past and future timesteps, e.g.\ $p_{\textrm{\textsubscript{IDM}}}(a_t | o_t, o_{t+1})$, and (2) a behavioral cloning (BC) model on trajectories of observations labeled with the IDM.
Data to train the IDM is collected by rolling out the BC model in the target environment such that both models improve in tandem.
However, at any point in training if there are sequences in the dataset that the IDM performs poorly on, it requires that the BC model perform those or similar sequences in order for the IDM to improve and correctly label them.
Therefore, if the BC model does not explore efficiently, it could severely slow down learning.
In order to avoid this potential issue we opted for a simpler two-stage approach: we first train an IDM on a small number of labeled trajectories collected from human contractors (they play the game as would normally as we record their keypresses and mouse movements). Because human contractors reach most relevant parts of the state space, we can hold the IDM fixed throughout BC training.

Compared to most previous work in semi-supervised imitation learning, we experiment in the much more complex and open-ended environment of Minecraft. Minecraft is a voxel-based 3D video game that, due its popularity and wide variety of mechanics, has attracted a vast amount of RL research.\cite{guss2019minerl, tessler2017deep, kanitscheider2021multi, oh2016control, abel2016exploratory, arumugam2019deep, trott2019keeping, alaniz2018deep, udagawa2016fighting, shu2017hierarchical, oh2017zero, shi2022learning, matiisen2019teacher, patil2020align, skrynnik2020forgetful, lin2021juewu}
A large body of work focuses on small, custom-made Minecraft worlds with tasks such as navigation,\cite{arumugam2019deep, matiisen2019teacher} block placing,\cite{trott2019keeping, alaniz2018deep} instruction following,\cite{oh2017zero, shi2022learning} combat,\cite{udagawa2016fighting} and others.\cite{oh2016control, tessler2017deep, shu2017hierarchical}
Work operating in the massive, randomly generated environments of Minecraft itself has included hill climbing,\cite{abel2016exploratory} automated curriculum learning~\cite{kanitscheider2021multi} and, most closely related to the RL experiments presented in Sec.~\ref{section:RL_finetuning}, diamond mining.\cite{guss2019minerl, patil2020align, skrynnik2020forgetful, lin2021juewu}
However, to the best of our knowledge, there is no published work that operates in the full, unmodified human action space, which includes drag-and-drop inventory management and item crafting.
\section{Methods}
\label{section:method}

\begin{figure}[t]
\begin{center}
\includegraphics[width=\linewidth]{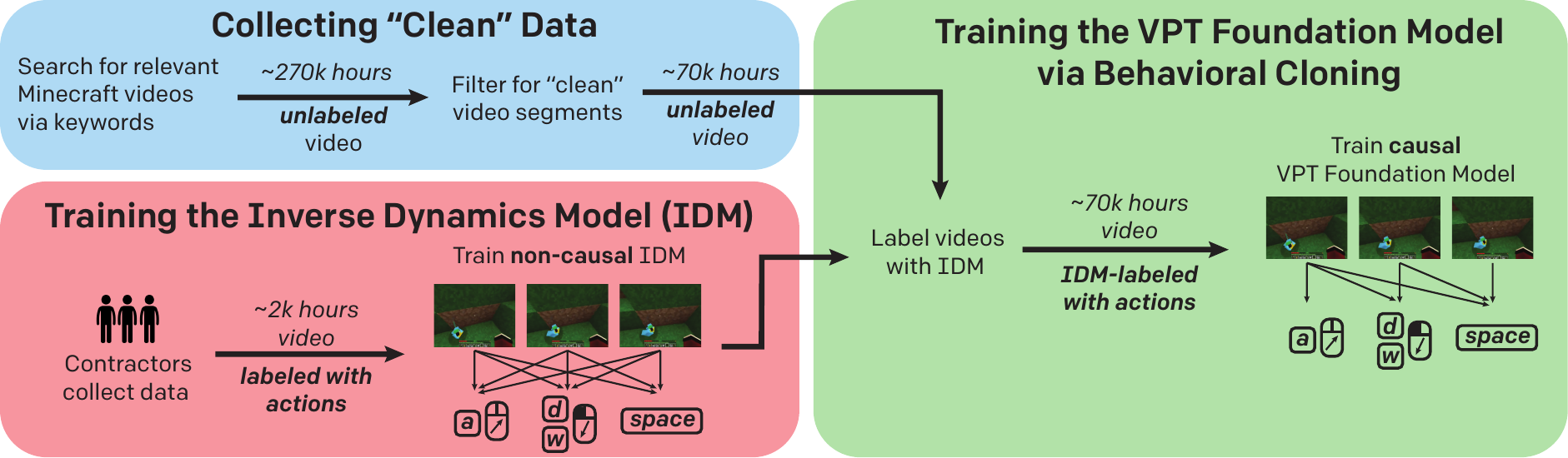}
\end{center}
\vspace{-10px}
\caption{Video Pretraining (VPT) Method Overview.}
\vspace{-10px}
\label{fig:method}
\end{figure}

\paragraph{Inverse Dynamics Models (IDM)} VPT, illustrated in Figure~\ref{fig:method}, requires we first collect a small amount of labeled contractor data with which to train an inverse dynamics model $p_{\textrm{\textsubscript{IDM}}}(a_t | o_{1...T})$, which seeks to minimize the negative log-likelihood of an action at timestep $t$ given a trajectory of $T$ observations $o_t~:~ t \in [1...T]$.
In contrast to an imitation learning policy, the IDM can be non-causal, meaning its prediction for $a_t$ can be a function of both past and \emph{future events}, i.e. $o_{t' > t}$.
Compared to the behavioral cloning objective of modeling the distribution of human intent given past frames only, we hypothesize that inverting environment dynamics is easier and more data efficient to learn.
Indeed, Sec.~\ref{section:results_IDM} will show that the IDM objective is much easier to learn, and furthermore Sec.~\ref{section:idm_data_scaling} will show that with very little labeled data (as few as 100 hours) we can train a fairly accurate IDM. This IDM can be used to label online videos, providing the large amount of data required for the harder task of behavioral cloning. See appendices ~\ref{appendix:training_details_IDM} and \ref{appendix:contractor_data} for IDM training and data collection details.

\paragraph{Data Filtering} We gather a large dataset of Minecraft videos by searching the web for related keywords (Appendix~\ref{appendix:collecting_internet_data}). Online videos often (1) include overlaid artifacts, such as a video feed of the player’s face, channel logos, watermarks, etc., (2) are collected from platforms other than a computer with different gameplay, or (3) are from different game modes, e.g.\ in Minecraft we only want "survival mode" where players start from scratch and must gather or craft all their items.
We call data ``clean'' if it does not contain visual artifacts and is from survival mode, and call all other data ``unclean.'' With enough data, a large enough model, and enough training compute, a BC model trained on both unclean and clean videos would likely still perform well in a clean Minecraft environment. However, for simplicity and training compute efficiency, we choose to filter out unclean segments of video (note that a video may contain both clean and unclean segments). We do this by training a model to filter out unclean segments using a small dataset (8800) of images sampled from online videos labeled by contractors as clean or unclean (Appendix~\ref{section:svm}).

\paragraph{VPT Foundation Model} We train a foundation model with standard behavioral cloning, i.e. minimizing the negative log-likelihood of actions predicted by the IDM on clean data. For a particular trajectory of length $T$ we minimize
\eqn{\min_{\theta}\sum_{t\in[1...T]}-\log\pi_{\theta}(a_t | o_1, \dots, o_t) \textrm{, where  }~ a_t\sim p_\textrm{\textsubscript{IDM}}(a_t | o_1, \dots, o_t, \dots, o_T)}
As we will see in the following sections, this model exhibits nontrivial zero-shot behavior and can be fine-tuned with both imitation learning and RL to perform even more complex skills. 

\section{Results}
\label{section:results}
\subsection{Performance of the Inverse Dynamics Model}
\label{section:results_IDM}

\begin{figure}[h]
\begin{center}
\includegraphics[width=\linewidth]{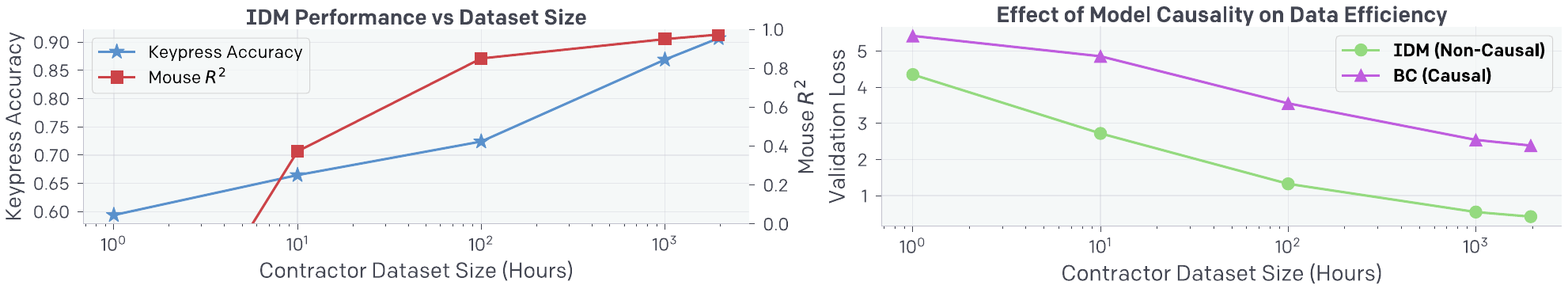}
\end{center}
\vspace{-8px}
\caption{\textbf{(Left)} IDM keypress accuracy and mouse movement $R^2$ (explained variance\cite{steel1960principles}) as a function of dataset size. \textbf{(Right)} IDM vs.\ behavioral cloning data efficiency.}
\label{fig:first_idm_results}
\end{figure}

The IDM architecture is comprised primarily of a temporal convolution layer, a ResNet\cite{he2016deep} image processing stack, and residual unmasked attention layers, from which the IDM simultaneously predicts keypresses and mouse movements (see Appendix \ref{appendix:training_details_IDM} for IDM architecture and training details).
A key hypothesis behind our work is that IDMs can be trained with a relatively small amount of labeled data.
While more data improves both mouse movement and keypress predictions, our best IDM trains on only 1962 hours of data (compared to the $\sim$70k hours of clean data we collected from the internet) and achieves 90.6\% keypress accuracy and a 0.97 $R^2$ for mouse movements evaluated on a held-out validation set of contractor-labeled data (Figure \ref{fig:first_idm_results} left).

Figure~\ref{fig:first_idm_results} (right) validates our hypothesis that IDMs are far more data efficient than BC models, likely because inverting environment mechanics is far easier than modeling the entire distribution of human behavior. The IDM is two orders of magnitude more data efficient than a BC model trained on the same data and improves more quickly with more data.
This evidence supports the hypothesis that it is more effective to use contractor data within the VPT pipeline by training an IDM than it is to train a foundation model from contractor data directly (Sections \ref{section:foundation_data_scaling} and \ref{section:idm_data_scaling} provide additional evidence).
\subsection{VPT Foundation Model Training and Zero-Shot Performance}
\label{section:results_foundation}

\begin{figure}[h]
    \centering
    \includegraphics[width=\linewidth]{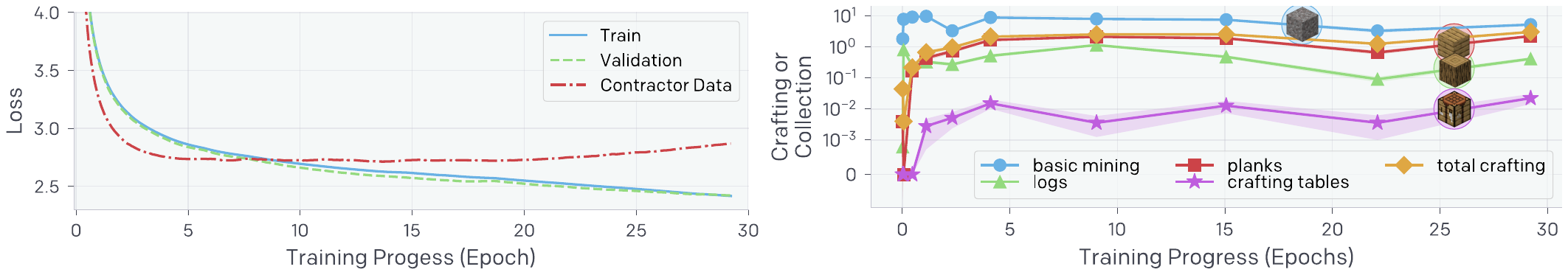} 
    \vspace{-6px}
    \caption{\label{fig:foundation} \textbf{(Left)} Training and validation loss on the \texttt{web\_clean} internet dataset with IDM pseudo-labels, and loss on the main IDM contractor dataset, which has ground-truth labels but is out-of-distribution (see text). \textbf{(Right)} Amount a given item was collected per episode averaged over 2500 60-minute survival episodes as a function of training epoch, shaded with the standard error of the mean. Basic mining refers to collection of dirt, gravel, or sand (all materials that can be gathered without tools). Logs are obtained by repeatedly hitting trees for three seconds, a difficult feat for an RL agent to achieve as we show in Sec.~\ref{section:RL_finetuning}. Planks can be crafted from logs, and crafting tables crafted from planks. Crafting requires using in-game crafting GUIs, and proficient humans take a median of 50 seconds (970 consecutive actions) to make a crafting table.
    }
\end{figure}

We now explore the emergent behavior learned by a behavioral cloning policy trained on an extremely large, but noisy, internet dataset labeled with our IDM.
To collect the unlabeled internet dataset, we searched for publicly available videos of Minecraft play with search terms such as ``minecraft survival for beginners.''
These searches resulted in $\sim$270k hours of video, which we filtered down to ``clean'' video segments yielding an \emph{unlabeled} dataset of $\sim$70k hours, which we refer to as \texttt{web\_clean} (Appendix~\ref{appendix:collecting_internet_data} has further details on data scraping and filtering).
We then generated pseudo-labels for \texttt{web\_clean} with our best IDM (Section~\ref{fig:first_idm_results}) and then trained the VPT foundation model with behavioral cloning. Preliminary experiments suggested that our model could benefit from 30 epochs of training and that a 0.5 billion parameter model was required to stay in the efficient learning regime \cite{kaplan2020scaling} for that training duration (Appendix~\ref{Appendix:FoundationModelScaling}), which took $\sim$9 days on 720 V100 GPUs.

We evaluate our models by measuring validation loss (Fig.~\ref{fig:foundation}, left) and rolling them out in the Minecraft environment. Unless otherwise noted, in all environment evaluations we spawn agents in a standard survival mode game where they play for 60 minutes, i.e. 72000 consecutive actions, and we plot the mean and shade the standard error of the mean for various game statistics such as crafting and collection rates (Fig.~\ref{fig:foundation}, right).
The VPT foundation model quickly learns to chop down trees to collect logs, a task we found near impossible for an RL agent to achieve with the native human interface (Sec.~\ref{section:RL_finetuning}).
It also learns to craft those logs into wooden planks and then use those planks to craft a crafting table, which are required to unlock most other technology in the game and take a human proficient in Minecraft approximately 50 seconds (970 consecutive actions) to collect.
While these behaviors are fairly complex in the native human action space, the VPT foundation model crafts these items at a rate far below that of our proficient contractors, e.g.\ on average our contractors craft 5.44 crafting tables in 60 minutes of play versus 0.19 for the foundation model.
The model also crafts a non-negligible amount of wooden sticks, which are required to make wooden tools; collects various flowers and crafts dyes from them; kills zombies that appear during the night; hunts wild animals; collects various berries and mushrooms and eats them; and finds game-generated villages from which to collect various rare items from chests. The model also learned to navigate uneven terrain, swim, and pillar jump, which involves the agent repeatedly jumping and quickly placing a block below itself such that it climbs upward by making a pillar.\footnote{Sample videos: \href{https://www.youtube.com/playlist?list=PLNAOIb\_agjf3U3rSvG\_BCWqJ869NdBhcP}{https://www.youtube.com/playlist?list=PLNAOIb\_agjf3U3rSvG\_BCWqJ869NdBhcP}}

While training and validation loss decrease healthily over training (Fig.~\ref{fig:foundation}, left), loss on our contractor dataset (which the VPT model does not train on) begins increasing after 7 epochs.
Contractor data could be out-of-distribution because our contractors may have a different distribution of play or because there is some impactful visual domain shift compared to videos from the web.
While one could have expected this would be predictive of declining evaluation performance, we do not see notable game statistics from the VPT foundation model rollouts (Figure \ref{fig:foundation}, right) decrease over training, and in the next section we show that BC fine-tuning performance continually improves as the VPT foundation model trains.
We provide more insight into this curious phenomenon in Appendix~\ref{Appendix:FoundationModelScaling}.

\subsection{Fine-Tuning with Behavioral Cloning}
\label{section:BC_finetuning}
Foundation models are designed to have a broad behavior profile and be generally capable across a wide variety of tasks.
To incorporate new knowledge or allow them to specialize on a narrower task distribution, it is common practice to fine-tune these models to smaller, more specific datasets.\cite{brown2020language}
The VPT foundation model trained on the broad \texttt{web\_clean} dataset had nontrivial zero-shot performance; it was able to craft a crafting table yet unable to go past this in the technology tree.
As a case study into BC fine-tuning, we attempt to improve the VPT foundation model's ability to collect and craft these ``early game'' items by fine-tuning to two narrower datasets targeted at Minecraft behavior within the first few minutes of players starting in a fresh world.
In the first dataset, \texttt{contractor\_house}, contractors have 10 minutes to build a basic house from scratch using primarily wood, sand, and dirt.
Collecting contractor data can be difficult and expensive, so we also construct a dataset \texttt{earlygame\_keyword} by searching for videos online with descriptions that match keywords such as “new world”, “let's play episode 1”, etc.; this is a subset of \texttt{web\_clean} and is labeled with the IDM. See Appendix~\ref{appendix:contractor_house_data} and~\ref{appendix:early_game_data} for full descriptions of both datasets.

\begin{figure}[h]
\begin{center}
\includegraphics[width=\linewidth]{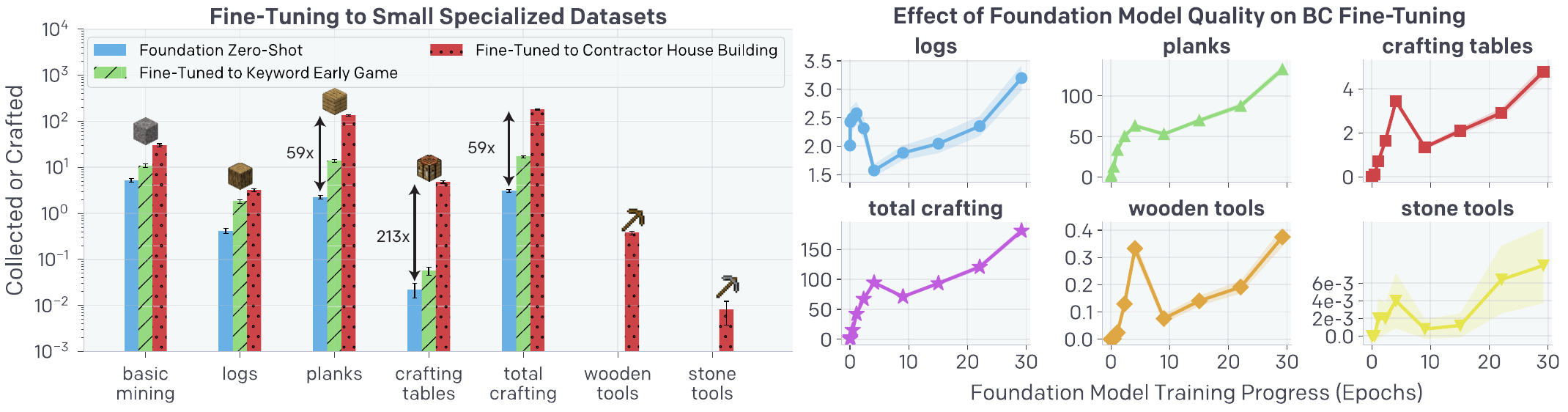}
\end{center}
\vspace{-8px}
\caption{\textbf{(Left)} Collection and crafting rates for three policies: the zero-shot VPT foundation model, and the VPT foundation model BC fine-tuned to the \texttt{earlygame\_keyword} or \texttt{contractor\_house} datasets. BC fine-tuning to either dataset improves performance, including (for the \texttt{contractor\_house} dataset) yielding wooden and stone tools.
Proficient Minecraft players take a median of 1.2 minutes (1390 actions) to construct wooden tools and 2.3 minutes (2790 actions) to construct stone tools. \textbf{(Right)} Collection and crafting rates for VPT foundation model snapshots throughout training \emph{after} they are BC fine-tuned to the \texttt{contractor\_house} dataset. In general, crafting-related behaviors increase throughout foundation model training. Fig.~\ref{fig:foundation} defines the other task terms (logs, planks, crafting tables, and total crafting).
}
\label{fig:main_finetuning}
\end{figure}

Fine-tuning to \texttt{earlygame\_keyword} results in a large boost compared to the zero-shot foundation model: 2.5x more crafting tables, 6.1x more planks, 4.3x more logs, and 5.5x more crafting overall (Fig.~\ref{fig:main_finetuning}).
However, when fine-tuning to this dataset we did not see any new behaviors emerge, only a refinement of existing skills. We saw an even bigger improvement when fine-tuning to the \texttt{contractor\_house} dataset: 213x more crafting tables, 59x more wooden planks, 7x more logs, and 59x more crafting over all. In addition, we saw the emergence of crafting wooden tools, which requires placing a crafting table on the ground, opening it to reveal a new crafting interface, and then using it to craft wooden tools. This entire sequence takes a proficient human player a median of 1.2 minutes (1390 consecutive actions) to accomplish. The model goes further and collects cobblestone, which requires a wooden pickaxe to mine, and crafts stone tools, requiring it to again use a crafting table; this takes a proficient human player a median of 2.3 minutes (2790 consecutive actions). We also saw this model more frequently raiding villages that randomly spawn in the game, hunting animals for food, in addition to many behaviors we saw performed by the foundation model.\footnote{Sample Videos: \href{https://www.youtube.com/playlist?list=PLNAOIb\_agjf2yDSs4AqcoyPv4z\_eWUiKm}{https://www.youtube.com/playlist?list=PLNAOIb\_agjf2yDSs4AqcoyPv4z\_eWUiKm}}

Despite the foundation model's zero-shot rollout performance plateauing 1/3 into training (Fig.~\ref{fig:foundation}, right), fine-tuning performance \textit{does} continue to increase throughout foundation model training (Fig.~\ref{fig:main_finetuning}, right).
Additionally, there is a stark difference in performance when training from scratch vs.\ fine-tuning from the VPT foundation model (Fig.~\ref{fig:main_finetuning} right, comparing the left and rightmost points).
\subsection{Fine-Tuning with Reinforcement Learning}
\label{section:RL_finetuning}
\begin{figure}[h]
\vspace{-8px}
\begin{center}
\includegraphics[width=\linewidth]{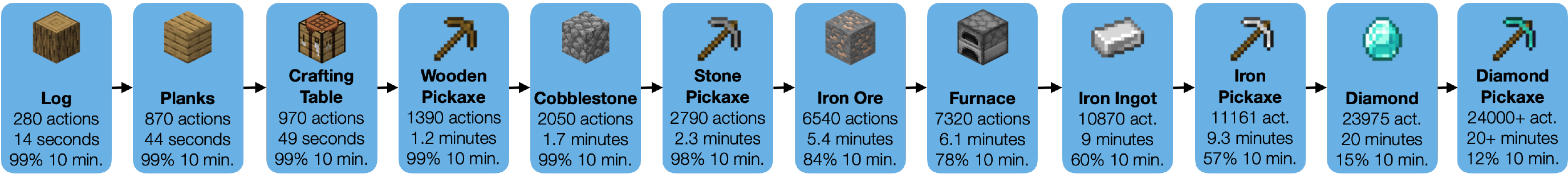}
\end{center}
\vspace{-8px}
\caption{
Typical sequence of items for obtaining a diamond pickaxe. 
Below each item is the median time and number of actions contractors required to obtain that item and the percentage of contractors that got the item within 10 minutes. The median time to obtain a diamond pickaxe is unknown (except that it is $>20$m) because contractors obtained this item in less than $50\%$ of 20-minute episodes.}
\label{fig:rlft_curriculum}
\end{figure}

To demonstrate the efficacy of RL fine-tuning, we chose the challenging goal of obtaining a diamond pickaxe within 10 minutes starting from a fresh Minecraft survival world.
Doing so involves acquiring a sequence of difficult-to-obtain items that require complex skills like mining, inventory management, crafting with and without a crafting table, tool use, operating a furnace, and mining at the lowest depths, where many hazards like enemies and lava exist (Fig.~\ref{fig:rlft_curriculum}). Adding to the difficulty, progress can be easily lost by dropping items, destroying items, or dying.
Obtaining a diamond pickaxe more often than not takes a proficient human over 20 minutes (24,000 actions).

Agents are rewarded for each item obtained in the sequence, with 
%reward caps for all items except diamonds and diamond pickaxes, and 
lower rewards for items that have to be collected in bulk and higher rewards for items near the end of the sequence. Agents are optimized with the phasic policy gradient~\cite{pmlr-v139-cobbe21a} RL algorithm for $\sim$1.3 million episodes (roughly $1.4\times 10^{10}$ frames). Episodes last for 10 minutes. See Appendix~\ref{Appendix:RLTrainingDetails} for reward function and RL training details. Due to computational constraints, RL experiments use a $\sim248$ million parameter VPT model (Appendix \ref{Appendix:FoundationModelScaling}).

A major problem when fine-tuning with RL is catastrophic forgetting \cite{kudithipudi2022biological, kirkpatrick2017overcoming} because previously learned skills can be lost before their value is realized.
For instance, while our VPT foundation model never exhibits the entire sequence of behaviors required to smelt iron zero-shot, it \textit{did} train on examples of players smelting with furnaces. It therefore may have some latent ability to smelt iron once the many prerequisites to do so have been performed.
To combat the catastrophic forgetting of latent skills such that they can continually improve exploration throughout RL fine-tuning, we add an auxiliary Kullback-Leibler (KL) divergence loss between the RL model and the frozen pretrained policy.\cite{vinyals2019grandmaster}

\begin{figure}[t]
\begin{center}
\begin{overpic}[width=0.49\linewidth]{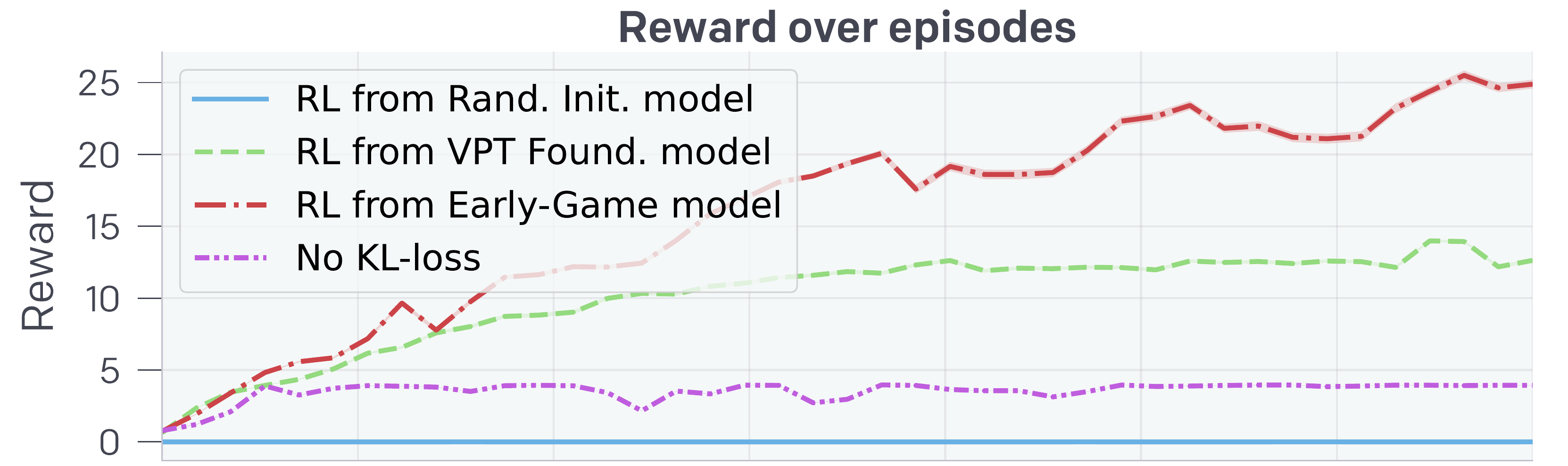}
\put(0, 27){\tiny(\textbf{a})}
\end{overpic}
\begin{overpic}[width=0.49\linewidth]{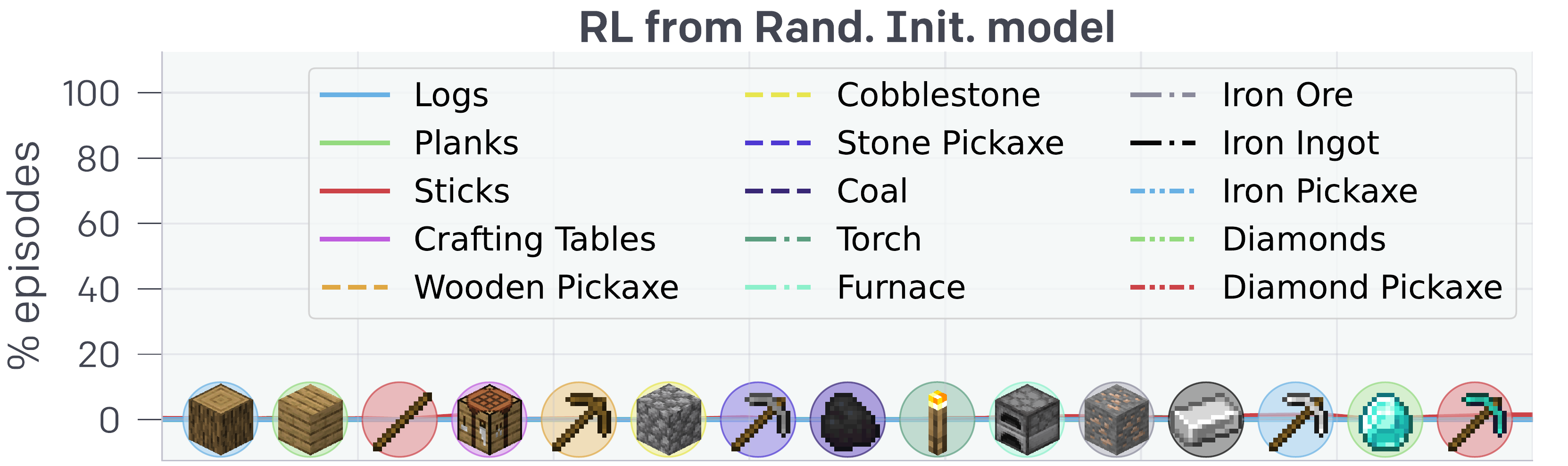}
\put(0, 27){\tiny(\textbf{b})}
\end{overpic}
\begin{overpic}[width=0.49\linewidth]{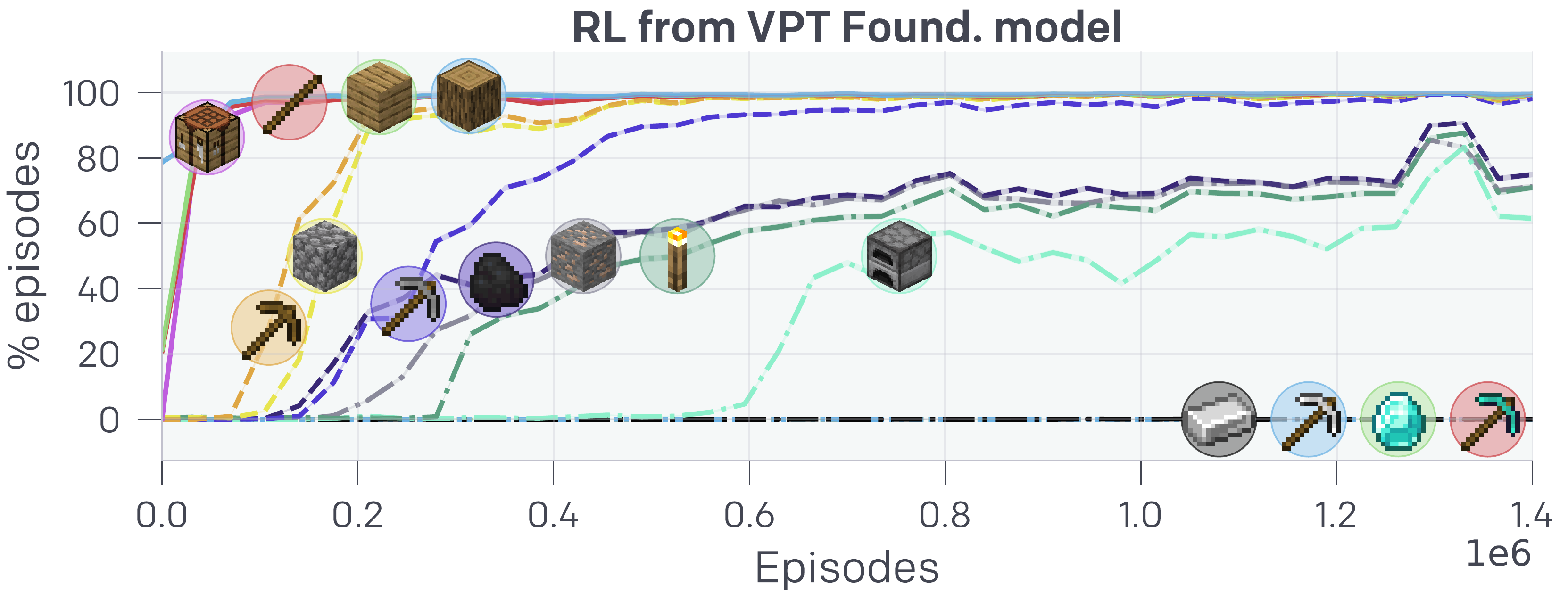}
\put(0, 36){\tiny(\textbf{c})}
\end{overpic}
\begin{overpic}[width=0.49\linewidth]{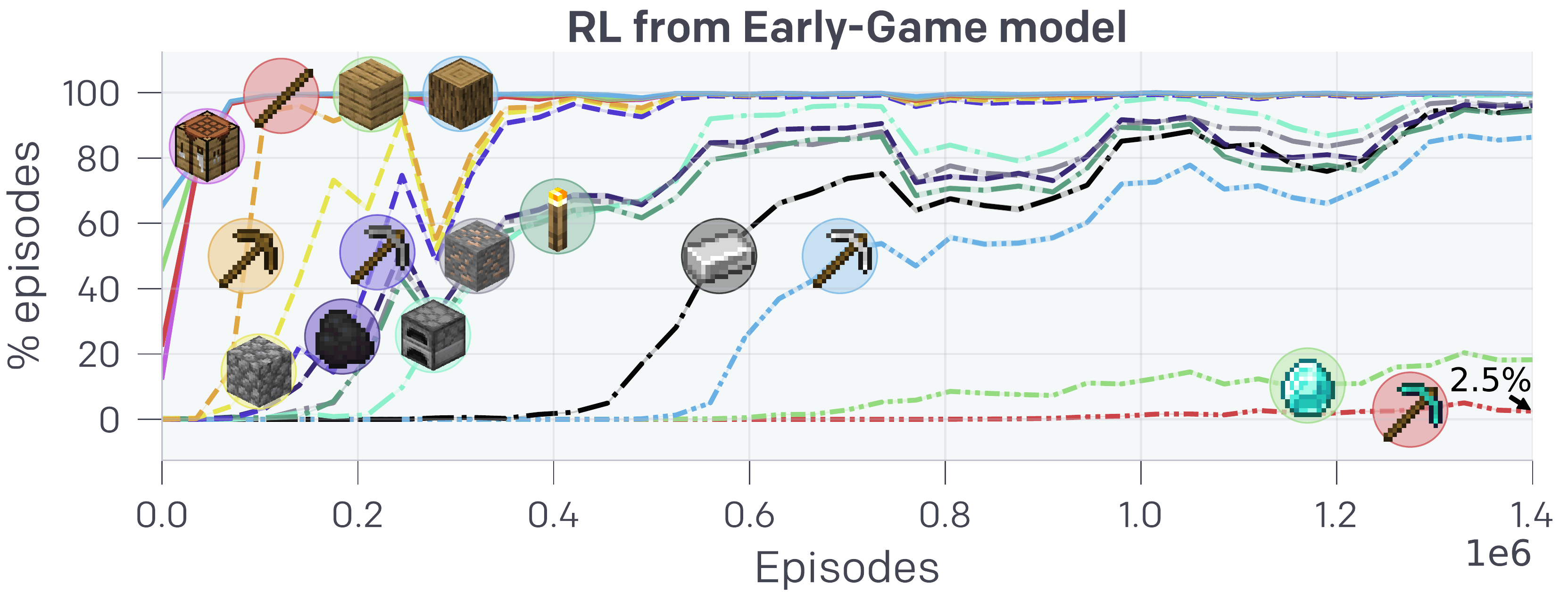}
\put(0, 36){\tiny(\textbf{d})}
\end{overpic}
\end{center}
\vspace{-8px}
\caption{
% Brief text
% (\textbf{a}) Reward over training for (1) RL from a randomly initialized policy and no KL loss, (2) RL fine-tuning from the VPT foundation model (Sec.~\ref{section:results_foundation}), (3) RL fine-tuning from from the VPT foundation model first BC fine-tuned to the \texttt{earlygame\_keyword} dataset (Sec.~\ref{section:BC_finetuning}), and (4) the same treatment as (3) but with no KL loss. (\textbf{b-d}) Per-episode success rate over the course of training of obtaining each item the agent is rewarded for collecting. \jc{some good info was lost in this caption rewrite, including that RL from scratch learns nearly nothing, and that from early game has 1.5\% on diamond pickaxe. We should add that in (and in general, the takeaway messages)}\bba{Those things were just repeating what is in the text, you want to say it twice? -- though if we have space i agree seems good to add in}
% Long text
RL Fine-tuning results. \textbf{(a)} RL from a randomly initialized model fails to get almost any reward, RL fine-tuning from the VPT foundation model performs substantially better with a reward near 13, and RL fine-tuning from the early-game model performs best with a reward of 25.
When training the early-game model without a KL loss to the original policy (\emph{No KL-loss}) progress stalls after 100,000 episodes, suggesting that the skills necessary to make further progress have been catastrophically forgotten.
\textbf{(b)} RL from a randomly initialized model occasionally collects sticks by breaking leaves (an easy but inefficient method of getting sticks that does not require logs or planks) and never learns to reliably collect logs.
\textbf{(c)} RL fine-tuning from the VPT Foundation model learns everything in the curriculum up to iron ore and making furnaces, but fails to learn to use the furnace to smelt iron ingots.
\textbf{(d)} RL fine-tuning from the early-game model learns to obtain (at human-level) all items in the sequence towards a diamond pickaxe and crafts a diamond pickaxe in $2.5\%$ of episodes.
}
\vspace{-8px}
\label{fig:rlft_items}
\end{figure}

Training from a randomly initialized policy fails to achieve almost \emph{any} reward, underscoring how hard an exploration challenge the diamond pickaxe task is for RL in the native human action space (Fig.~\ref{fig:rlft_items}a).
The model never learns to reliably collect logs, typically the first of many steps to obtaining a diamond pickaxe (Fig.~\ref{fig:rlft_items}b).
RL fine-tuning from the VPT foundation model does substantially better (Fig.~\ref{fig:rlft_items}a), learning everything up to mining iron ore and crafting furnaces. (Fig.~\ref{fig:rlft_items}c). However, this agent fails at smelting an iron ingot, the next item required to get further into the tech tree, likely because the zero-shot probability that the VPT foundation model smelts an iron ingot is too low, even when given the prerequisite materials.

Results further improve by first BC fine-tuning the VPT Foundation Model to the \texttt{earlygame\_keyword} dataset (the \emph{early-game model}, Sec.~\ref{section:BC_finetuning}) and then fine-tuning with RL (Fig.~\ref{fig:rlft_items}a), which in preliminary experiments we found to perform better than first fine-tuning to \texttt{contractor\_house} followed by fine-tuning with RL (Appendix~\ref{Appendix:RLPreliminary}).
The three-phase training (pretraining, BC fine-tuning, and then RL fine-tuning) succeeds in learning extremely difficult tasks: it achieves over $80\%$ reliability on iron pickaxes, almost $20\%$ reliability on collecting diamonds, and $2.5\%$ reliability on obtaining a diamond pickaxe (Fig.~\ref{fig:rlft_items}d). 
For comparison, human players given the objective of obtaining a diamond pickaxe collect these items in $57\%$, $15\%$, and $12\%$ of episodes, respectively, meaning our model is human-level for crafting iron pickaxes and mining diamonds.
Others have managed to obtain diamonds with $\sim0.1\%$ reliability in 15 minutes\cite{skrynnik2020forgetful, patil2020align} but always with a simplified action space designed to ease exploration. To the best of our knowledge, \textbf{we are the first to report non-zero success rates on crafting a diamond pickaxe}. Qualitatively, the model developed useful skills for diamond mining, such as efficient mining patterns, cave exploration, returning to previously placed objects like crafting tables, and advanced techniques like using wooden pickaxes as fuel when moving on to iron tools.\footnote{Videos found at \href{https://www.youtube.com/playlist?list=PLNAOIb\_agjf3e\_UKweM5pQUSfTw8r-Wfc}{https://www.youtube.com/playlist?list=PLNAOIb\_agjf3e\_UKweM5pQUSfTw8r-Wfc}}

Finally, we validated the importance of the KL loss to the pretrained model during RL fine-tuning. The treatment without a KL loss obtains only items early in the sequence (logs, planks, sticks, and crafting tables) limiting its reward (Fig.~\ref{fig:rlft_items}a).
This failure to progress further into the sequence is likely because, while the initial skills of chopping logs and crafting planks are being learned with RL, subsequent skills like crafting a wooden pickaxe are lost due to catastrophic forgetting.

\subsection{Data Scaling Properties of the Foundation Model}
\label{section:foundation_data_scaling}

\begin{figure}[h]
\begin{center}
\includegraphics[width=\linewidth]{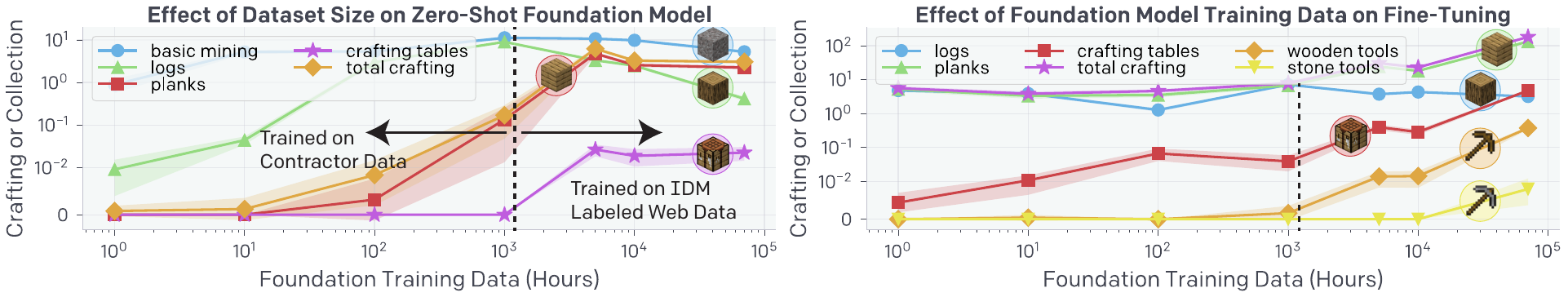}
\end{center}
\vspace{-10px}
\caption{\textbf{(Left)} Zero-shot rollout performance of foundation models trained on varying amounts of data. Models to the left of the dashed black line (points $\le$1k hours) were trained on contractor data (ground-truth labels), and models to the right were trained on IDM pseudo-labeled subsets of \texttt{web\_clean}. Due to compute limitations, this analysis was performed with smaller (71 million parameter) models except for the final point, which is the 0.5 billion parameter VPT foundation model. \textbf{(Right)} The corresponding performance of each model \emph{after} BC fine-tuning each model to the \texttt{contractor\_house} dataset.
}
\label{fig:foundation_data_scale}
\end{figure}

In this section we validate a core hypothesis behind this work: that it is far more effective to use labeled contractor data to train an IDM within the VPT method than it is to directly train a BC foundation model from that same small contractor dataset.
If we could cheaply collect a labeled contractor dataset of a similar order of magnitude as \texttt{web\_clean}, then this would not be important; however, collecting that scale of data would have cost millions of dollars.
Figure~\ref{fig:foundation_data_scale} compares foundation models trained on increasing orders of magnitude of data from 1 hour up to the full $\sim$70k \texttt{web\_clean} dataset.
Foundation models trained up to and including 1k hours are trained on the IDM contractor data, and those trained on 5k hours and above are trained on subsets of \texttt{web\_clean}, which does not contain any IDM contractor data.
Scaling training data increases log collection, mining, and crafting capabilities.
The zero-shot model only begins to start crafting crafting tables at over 5000 hours of training data.
When fine-tuning each foundation model to \texttt{contractor\_house}, we see that crafting rates for crafting tables and wooden tools increase by orders of magnitude when using the entire $\sim$70k hour \texttt{web\_clean} dataset. We furthermore only see the emergence of crafting stone tools at the largest data scale.

\subsection{Effect of Inverse Dynamics Model Quality on Behavioral Cloning}
\label{section:idm_data_scaling}

\begin{wrapfigure}{r}{0.5\textwidth}
\vspace{-20px}
\begin{center}
\includegraphics[width=\linewidth]{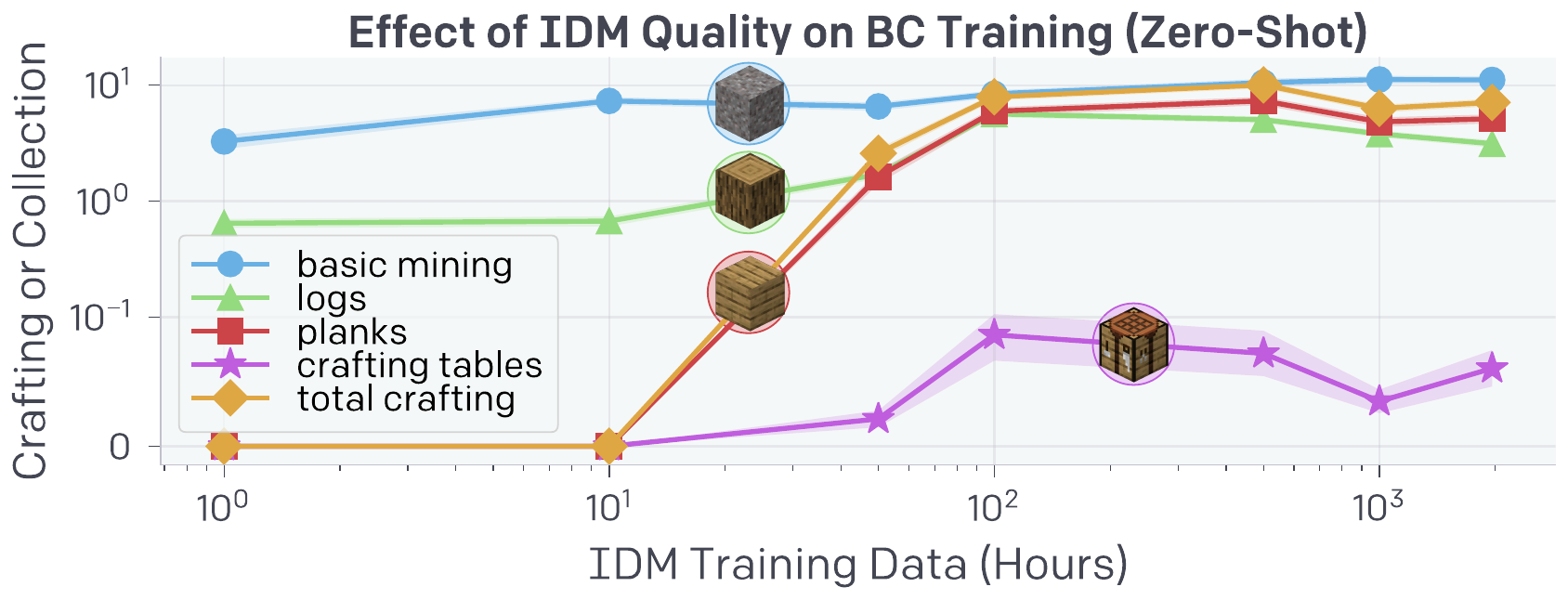}
\end{center}
\vspace{-8px}
\caption{Zero-shot performance of BC models trained from scratch on the \texttt{earlygame\_keyword} dataset labeled with IDMs that were trained on increasing amounts of contractor data.
}
\label{fig:idm_data_effect}
\vspace{-10px}
\end{wrapfigure}

This section investigates how downstream BC performance is affected by IDM quality. We train IDMs on increasingly larger datasets and use each to independently label the \texttt{earlygame\_keyword} dataset (this smaller dataset was chosen due to a limited compute budget). We then train a BC model from scratch on each dataset and report game statistics for each model as a function of IDM contractor dataset size (Fig.~\ref{fig:idm_data_effect}).

IDMs trained on at least 10 hours of data are required for any crafting, and the crafting rate increases quickly up until 100 hours of data, after which there are few to no gains and differences are likely due to noise. Similarly, crafting tables are only crafted after 50 or more hours of IDM data, and again gains plateau after 100 hours. While in all previous experiments we use our best IDM trained on 1962 hours of data, these results suggest we could reduce that number to as low as 100 hours.
\section{Discussion and Conclusion}
\label{section:discussion}
The results presented in this paper help pave the path to utilizing the wealth of unlabeled data on the web for sequential decision domains. 
Compared to generative video modeling or contrastive methods that would only yield representational priors, 
VPT offers the exciting possibility of directly \emph{learning to act} during pretraining and using these learned behavioral priors as extremely effective exploration priors for RL.
VPT could even be a better general representation learning method even when the downstream task is not learning to act in that domain—for example, fine-tuning to explain what is happening in a video—because arguably the most important information in any given scene would be present in features trained to correctly predict the distribution over future human actions. We leave this intriguing direction to future work.

Future work could improve results with more data (we estimate we could collect >1M hours) and larger, better-tuned models. Furthermore, all the models in this work condition on past observations only; we cannot ask the model to perform specific tasks. Appendix~\ref{appendix:text_conditioning} presents preliminary experiments on conditioning our models on closed captions (text transcripts of speech in videos), showing they become weakly steerable; we believe this a rich direction for future research.
Also, loss was not consistently correlated with downstream evaluation metrics (Sec.~\ref{section:results_foundation}), which often made progress slow and hard-won.
Another fruitful future direction would be to investigate the correlation between various training metrics and downstream evaluations.
Finally, while we do not anticipate any direct negative societal impacts from the models trained in this work, as VPT improves and expands to other domains it will be important to assess and mitigate harms that emerge with other forms of pretraining on internet datasets, such as emulating inappropriate behavior.\cite{bender2021dangers} 

In conclusion, VPT extends the paradigm of training large and general purpose behavioral priors from freely available internet-scale data to sequential decision domains.
Our models exhibited impressive zero-shot behavior and, when fine-tuned with RL, achieved an unprecedented result of crafting a diamond pickaxe in Minecraft (all the more difficult given the human interface).
We further showed that contractor data is far better used within the VPT pipeline than to train a foundation model directly and that only a small amount of contractor data (about \$2000 USD) was required to unlock massive amounts of unlabeled online data for use in BC. 
Finally, learning with the human keyboard and mouse interface is highly general and allows losslessly modeling the entire distribution of human behavior.
While we only experiment in Minecraft, we believe that VPT provides a general recipe for training behavioral priors in hard, yet generic, action spaces in any domain that has a large amount of freely available unlabeled data, such as computer usage.

%%%%%%%%%%%%%%%%%%%%%%%%%%%%%%%%%%%%%%%%%%%%%%%%%%%%%%%%%%%%

\bibliography{references}{}
\bibliographystyle{unsrtnat}

\section*{Acknowledgements} %commenting out so we do not accidentally forget and deanonymize during submission

 We thank the following people for helpful discussions and support: Bob McGrew, Ken Stanley, Joel Lehman,
 Ilya Sutskever, 
 Wojciech Zaremba, 
 Ingmar Kanitscheider, 
 David Farhi,
 Glenn Powell,
 Jonathan Gordon, and the OpenAI supercomputing team, especially Christian Gibson, Ben Chess, and Christopher Berner.

\newpage
\begin{center}
\textbf{\Large{Supplementary Information}}    
\end{center}
\appendix
\section{Collecting Internet Data}
\label{appendix:collecting_internet_data}
\subsection{Initial Unclean Dataset Curation}\label{section:data_collection}

Our goal was to curate a video dataset of Minecraft gameplay from the survival game mode. Additionally, we prefer the data come from game modes as close as possible to our evaluation environment, meaning preferably coming from Minecraft version 1.16, being on a computer (which uses a mouse and keyboard vs. video game controllers with keypads and other buttons), being single- (vs. multi-) player, and having the default look of the game (vs. modifications that alter that style, such as to make it look realistic). To try to accomplish these goals, we collect a dataset by performing keyword searches of publicly available videos on the internet. A list of search queries we used are given in Table~\ref{tab:query_strings}.

\begin{table}[h]
\centering
\begin{center}
\begin{tabular}{ |l|}
\hline
minecraft survival longplay \\
minecraft gameplay no webcam \\
minecraft gameplay survival mode \\
minecraft survival tutorial \\
minecraft survival guide \\
minecraft survival let's play \\
minecraft survival for beginners \\
minecraft beginners guide \\
ultimate minecraft starter guide \\
minecraft survival guide 1.16 \\
minecraft how to start a new survival world \\
minecraft survival fresh start \\
minecraft survival let's play episode 1 \\
let's play minecraft episode 1 \\
minecraft survival 101 \\
minecraft survival learning to play \\
how to play minecraft survival \\
how to play minecraft \\
minecraft survival basic \\
minecraft survival for noobs \\
minecraft survival for dummies \\
how to play minecraft for beginners \\
minecraft survival tutorial series \\
minecraft survival new world \\
minecraft survival a new beginning \\
minecraft survival episodio 1 \\
minecraft survival {\sffamily\foreignlanguage{russian}{epizod}} 1 \\
minecraft survival 1. bölüm \\
i made a new minecraft survival world \\
\hline
\end{tabular}
\end{center}
\caption{\label{tab:query_strings}Search terms used for generating the initial web dataset.}
\end{table}

For videos that have metadata available, we perform an additional step of metadata-based filtering to eliminate videos that do not fit our target distribution. In this step, we look for a list of blacklist keywords in the video title and description and reject videos that contain these terms. The blacklist keywords we use are: \{ps3, ps4, ps5, xbox 360, playstation, timelapse, multiplayer, minecraft pe, pocket edition, skyblock, realistic minecraft, how to install, how to download, realmcraft, animation\}. This process yielded us $\sim$270k hours of unlabeled data, which we filter down to only a ``clean'' subset as described in the next section.

\subsection{Training a Model to Filter out Unclean Video Segments}\label{section:svm}

We restrict the scope of this work to the Minecraft Survival game mode and therefore limit our training dataset to clips that are obtained from this mode that are relatively free from visual artifacts. To do so, we asked contractors to label a set of random video frames (images) from Minecraft videos (N=8800). These images were from  a random subset of the videos we collected toward the beginning of the project (Section~\ref{section:data_collection}).

\subsubsection{Label Collection}

\begin{figure}[htbp]
\includegraphics[width=\linewidth]{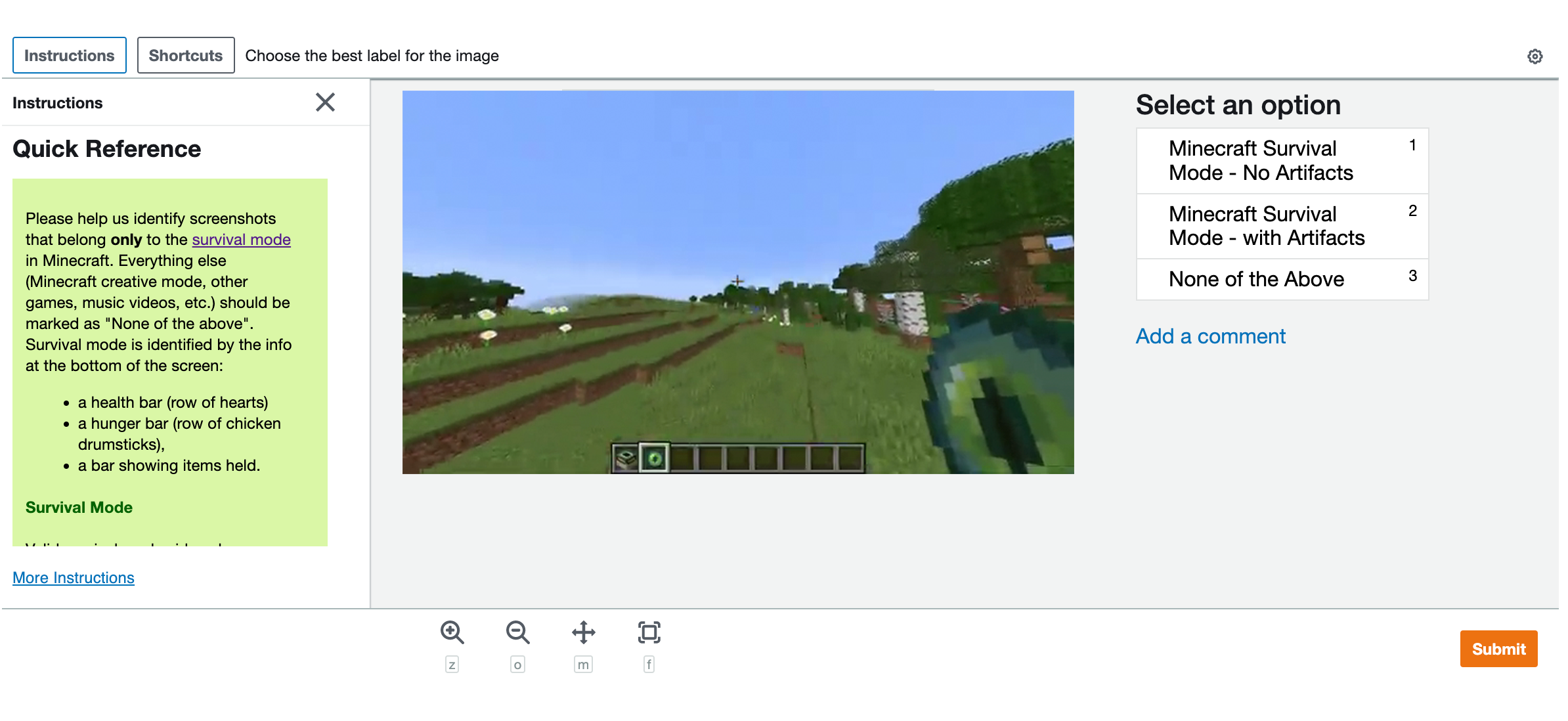} 
\caption{Amazon Mechanical Turk worker interface showing an example labeling task}
\label{fig:mTurk_worker_interface}
\end{figure}

We asked 5 workers on Amazon Mechanical Turk (mTurk) that we selected with a sample qualification task to label random screen capture images to be used in training the classifier. A sample worker interface that the workers saw on mTurk is given in Figure~\ref{fig:mTurk_worker_interface}.

\begin{figure}[htbp]
    \centering
     \begin{subfigure}{0.3\textwidth}
        \centering
        \includegraphics[width=\linewidth]{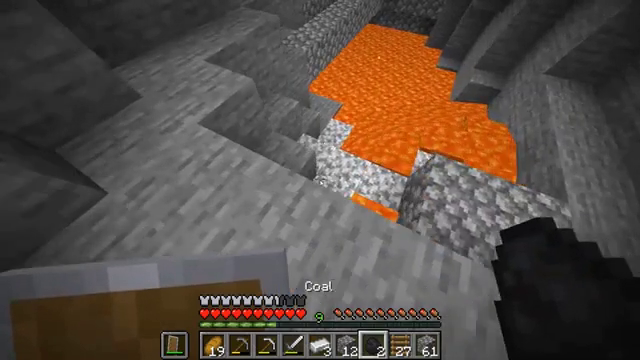} 
    \end{subfigure}
    \begin{subfigure}{0.3\textwidth}
        \centering
        \includegraphics[width=\linewidth]{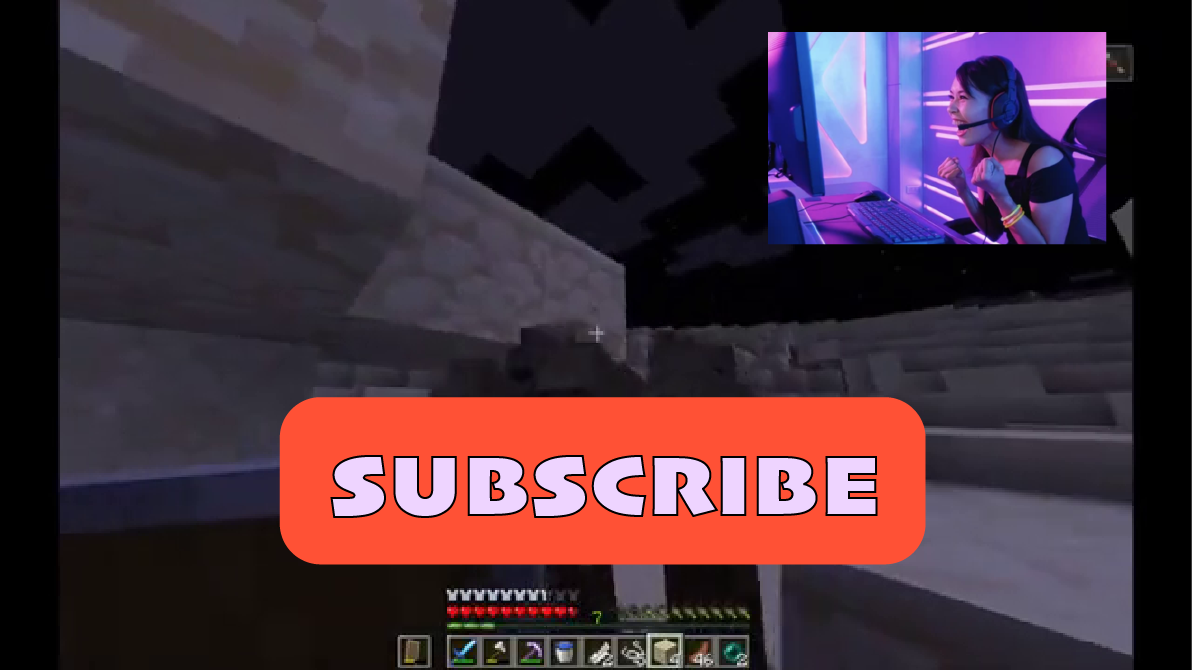} 
    \end{subfigure}
    \begin{subfigure}{0.3\textwidth}
        \centering
        \includegraphics[width=\linewidth]{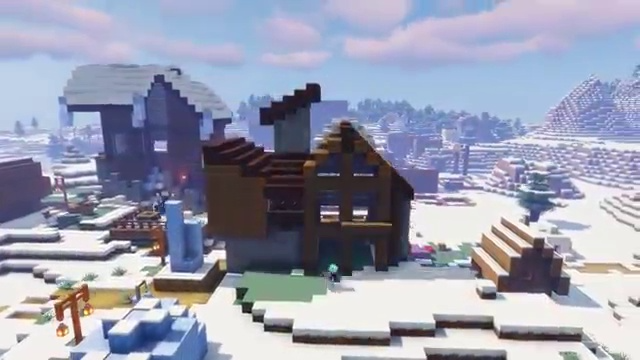} 
    \end{subfigure}
    \caption{\label{fig:a_1_data_class_examples}\textbf{(Left)} Sample image for Class 1: \texttt{Minecraft Survival Mode - No Artifacts}. \textbf{(Middle)} Sample image for Class 2: \texttt{Minecraft Survival Mode - with Artifacts} – Image contains annotations and picture-in-picture of the narrator. \textbf{(Right)} Sample image for Class 3: \texttt{None of the Above} – Image is missing the hotbar as well as health and armor bars, indicating that it was not captured during survival mode gameplay}
\end{figure}

We asked workers to label videos as being in one of the following three categories (see Figure~\ref{fig:a_1_data_class_examples} for visual examples of each class):
\begin{enumerate}
    \item \texttt{Minecraft Survival Mode - No Artifacts:} Video frames (images) that correspond to the Minecraft Survival game mode that do not contain any non-game visual artifacts (e.g.\ subscribe buttons, channel logos, advertisements, picture-in-picture of the narrator, etc.).
    \item \texttt{Minecraft Survival Mode - with Artifacts:} Video frames (images) of the Minecraft Survival game mode that include such visual artifacts. 
    \item \texttt{None of the Above:} Video frames (images) that are not from the Minecraft survival game mode, including those from other Minecraft game modes such as \emph{creative} mode or even other games/topics entirely.
\end{enumerate}

The full set of instructions workers received are as follows (note that we also included multiple image examples from each category in the worker instructions, similar to the sample subset provided in Figure~\ref{fig:a_1_data_class_examples}):

\noindent\fbox{%
    \parbox{\textwidth}{
~\\
Please help us identify screenshots that belong only to the survival mode in Minecraft. Everything else (Minecraft creative mode, other games, music videos, etc.) should be marked as \texttt{None of the above}. Survival mode is identified by the info at the bottom of the screen:
\begin{itemize}
    \item a health bar (row of hearts)
    \item a hunger bar (row of chicken drumsticks)
    \item a bar showing items held
\end{itemize}

{\bf Survival Mode}

Valid survival mode videos have health/hunger bars and an item hotbar at the bottom of the screen. 

{\bf Creative Mode}

Creative mode only has an item hotbar and should be classified as \texttt{None of the Above}. 

~\\

{\bf Label Descriptions}

\begin{itemize}
\item
\texttt{Minecraft Survival Mode - No Artifacts:}
These images will be clean screenshots from the Minecraft survival mode gameplay without any noticeable artifacts.
\item
\texttt{Minecraft Survival Mode - with Artifacts:}
These images will be valid survival mode screenshots, but with some added artifacts. Typical artifacts may include image overlays (a logo/brand), text annotations, a picture-in-picture of the player, etc.
\item
\texttt{None of the Above:}
Use this category when the image is not a valid Minecraft survival screenshot. It may be a non-Minecraft frame or from a different game mode. In non-survival game modes such as the creative mode, the health/hunger bars will be missing from the image, the item hotbar may or may not be still present.
\end{itemize}
}}

In total, we spent \$319.96 on human labeling experiments on mTurk, of which \$159.98 was directly paid to workers. The remaining amount was spent towards Amazon platform fees. The workers received \$0.01 per labeled image, at an hourly compensation of \$7.20 (based on an estimated labeling time of 5 seconds/image – in our internal sample run of the same task, we found the average labeling time to be < 3 seconds).

Since we perform rigorous keyword and metadata based filtering of videos (as described in \ref{section:data_collection}) from which we served sample images to be labeled, serving offensive content to workers was extremely low risk and no such images were detected during our manual checks. We only collected labels during our experiment, and the workers were fully anonymized via the mTurk platform, therefore no personally identifiable information (PII) was collected.

\subsubsection{SVM Training}

With the image labels collected as described in the previous section, we trained a classifier to extract video segments that consist of frames from the \texttt{Minecraft Survival Mode - No Artifacts} category.
Given a set of labeled images, we obtain embeddings for each image using the RN50x64 ResNet CLIP Model.\cite{radford2021learning} This is a ResNet-based CLIP model that is scaled up to have approximately 64x the compute of a ResNet-50. We then train a Support Vector Machine (SVM) using the RBF kernel to obtain a frame classifier. We use the Scikit-learn\cite{pedregosa2011scikit} SVM implementation with the parameter configuration given in Table~\ref{svmtable}.

\begin{table}[htbp]
\centering
\begin{center}
\begin{tabular}{ |l|l|l|}
\hline
 CLIP Model Specification & \multicolumn{2}{l|}{RN50x64 (see text)}\\
 \hline
 CLIP Input Image Resolution & \multicolumn{2}{l|}{448x448x3} \\
 \hline
 CLIP Embedding Feature Length & \multicolumn{2}{l|}{1024} \\
 \hline
 \multirow{3}{*}{SVM Parameters} & Kernel & \texttt{rbf} \\
 \cline{2-3} 
 & C & \texttt{20}\\
 \cline{2-3} 
 & Gamma & \texttt{scale} \\
  \hline
 \multirow{3}{*}{Sample Size} & Class 1 & 2200 \\
 \cline{2-3} 
 & Class 2 & 2200\\
 \cline{2-3} 
 & Class 3 & 4400 \\
 \hline
\end{tabular}
\end{center}
\caption{Feature Extraction Details and SVM Configuration. The parameters are for the SVM implementation in Scikit-learn\cite{pedregosa2011scikit}.
}
\label{svmtable}
\end{table}

Finally, we apply the classifier to frames of raw video sequences at a rate of 3 frames/second. We filter for videos that consist of at least 80\% "clean" frames at this stage (Classes \texttt{Minecraft Survival Mode - with Artifacts} and \texttt{None of the Above} are both considered not clean). From this set, we apply a median filter (with a kernel size of 7) to the labels and segment videos by splitting the "clean" segments that are at least 5s in duration. The result of this is our final \texttt{web\_clean} dataset.

\subsection{\texttt{early\_game} Dataset}
 \label{appendix:early_game_data}
The \texttt{early\_game} dataset is a $\sim$3000 hour subset of \texttt{web\_clean} targeted at ``early game'' Minecraft behavior, i.e. instances where players start in a fresh world with no items. We obtain the metadata text that accompanies the videos in \texttt{web\_clean} and determine whether any of the following regular expressions match:
\begin{itemize}
    \item (ep|episode|eps|day|session|sesh|chapter|chap\\.|series|part|parte|pt|round|day|t\d{\^{a}}p|b\"ol\"um|episodio|\sffamily\foreignlanguage{russian}{епизод}|\sffamily\foreignlanguage{russian}{эпизод})( )*(\textbackslash.1|\#1|1|\textbackslash.01|\#01|01|one[\hspace{1.5px}$\hat{ }$\hspace{1.5px}0-9]|\$)
    \item start
    \item beginning
    \item (new|fresh|clean).*(world|game|play)
    \item from scratch
\end{itemize}
From this set of videos, we take only the first 5 minutes of each video.
\section{Contractor Data}
\label{appendix:contractor_data}

\subsection{Recording Contractor Play}
Our contractors use a custom Minecraft recorder that we built that records their actions and game video feeds as they play. The recorder is implemented using the MCP-Reborn (\href{https://github.com/Hexeption/MCP-Reborn}{github.com/Hexeption/MCP-Reborn}) modding package. To ensure that the recorder environment is as close as possible to the Minecraft environment used for RL rollouts and evaluations (Appendix~\ref{appendix:minecraft_environment}), we use the same underlying game engine for both. The recorder is a Java app that runs in a window mode, with constant resolution of 1280x760. Brightness is set to 0 (the "gloomy" setting in Minecraft), which is the default setting. Other graphics settings (field of view, GUI scale) are fixed to the values used in the Minecraft environment (\ref{appendix:minecraft_env_observation_space}); we explicitly prevented users from changing graphics settings.
Unlike the environment, the recorder allows all keyboard key presses and continuous (as opposed to binned) mouse actions. On every game step (or ``tick'') the frame buffer used to display the game window is downsized to 640x360 and written into a video file. In-game actions are recorded in a separate JSONL file (a text file where each line is a JSON-formatted string). All recordings are chunked into 5 minute clips: after each 5 minute segment of contractor game play the recorder automatically uploads the video file, the JSONL file with actions, as well as a Minecraft state file. To ensure that contractors cannot corrupt each other's data, we provided every contractor with an individual cloud bucket, as well as with credentials giving write access only to that bucket. Credentials also included adjective-adjective-noun names (e.g.\ grumpy-amethyst-chipmunk), generated with the \texttt{namegenerator} python package to ensure contractor anonymity when we publish the data.

\subsection{Contractor Contract}
\label{appendix:contractor_contract}

We recruited contractors by posting the following offer on the UpWork freelancing platform. 

\begin{displayquote}
``We are collecting data for training AI models in Minecraft. You'll need to install java, download the modified version of Minecraft (that collects and uploads your play data), and play Minecraft survival mode! Paid per hour of gameplay. Prior experience in Minecraft not necessary. We do not collect any data that is unrelated to Minecraft from your computer.''
\end{displayquote}

We had the applications open for a day, and then randomly selected 10 applicants for the first round of contractors. Later in the project, as we needed more data and as some contractors asked to terminate their contracts, we added more applicants from the original pool as well as referrals from the currently working contractors.
The contractors were paid \$20 per hour (minus Upwork platform fees and applicable taxes). All of the results presented in this paper are based on about 4,500 hours of data (including data recorded to gather statistics of human play that was not used for training), which cost us around \$90,000. Over the course of the project, we collected some data we did not use due to bugs in the recorder and for some ideas we ultimately did not pursue. In total, we spent about \$160k for contractor compensation over the course of the project. However, as we discuss in Sec.~\ref{section:idm_data_scaling}, we could likely obtain most of our results with an IDM trained using only \$2000 worth of data, i.e. the foundation VPT model, BC fine-tuning to the \texttt{earlygame\_keyword} dataset, and the RL fine-tuning results. Collecting the \texttt{contractor\_house} dataset cost about \$8000. Because we used the IDM trained on about 2000 hours of contractor data, the actual cost of contractor data for those results was around \$40,000.

In early stages of the project, we were planning to use contractor data solely for the purpose of training the IDM.
As such, no specific tasks were given, other than ``play the survival mode of Minecraft like you normally would.''
Later in the project, we requested that contractors perform specific tasks in Minecraft, such as:
\begin{itemize}
    \item Collect as many units of wood as possible, using only wooden or stone tools (\texttt{treechop})
    \item Start a new world every 30 minutes of game play
    \item Build a basic house in 10 minutes using only dirt, wood, sand, and either wooden or stone tools (\texttt{contractor\_house}, more details below in Appendix~\ref{appendix:contractor_house_data}).
    \item Starting from a new world and an empty inventory, find resources and craft a diamond pickaxe in 20 minutes (\texttt{obtain\_diamond\_pickaxe}). This dataset was used to obtain statistics for how long it takes humans on average to complete this task (and the subtasks required to complete it) when obtaining a diamond pickaxe is their goal.
\end{itemize}

Since we only recorded in-game events and videos, the data does not include personally identifiable information. That being said, the contractors could theoretically use Minecraft's open-world property to generate personally identifiable information and/or offensive content (e.g.\ by using Minecraft blocks to write their name or offensive messages, then finding a spot from which the message would be visible). In practice, we have not seen any attempts to do so in the contractor videos that we watched. Of course, 
we train our BC models on videos from the internet of people playing Minecraft, and if such behavior is in those videos our model could also potentially learn it, although we expect such behavior is rare enough that our model would not be likely to reproduce it.

\subsection{Data for the Inverse Dynamics Model.}
Since the IDM's task is to infer actions given the video, any labelled data is appropriate for IDM training. In practice, 
we included general gameplay as well as the \texttt{treechop} task data described in the previous section, which amounted to a total of 1962 hours. Due to collecting datasets like \texttt{contractor\_house} only at late stages of the project, they were not included in IDM training.

\subsection{contractor\_house.}
\label{appendix:contractor_house_data}
The \texttt{contractor\_house} contains about 420 hours of data. We asked contractors to  build a basic house in 10 minutes, using only basic dirt, wood, and sand, blocks. Each trajectory starts in a newly generated world and a timer forcibly ends a trajectory after a 20 minute time limit. For this task, many contractors chose to begin their trajectories by crafting basic tools and building blocks, specifically it was common for the first 2 minutes to be spent crafting a wooden pickaxe and then mining stone for an assortment of stone tools before gathering more building blocks and beginning to create their structure.
\begin{figure}[t]
    \centering
     \begin{subfigure}{0.358\textwidth}
        \centering
        \includegraphics[width=\linewidth]{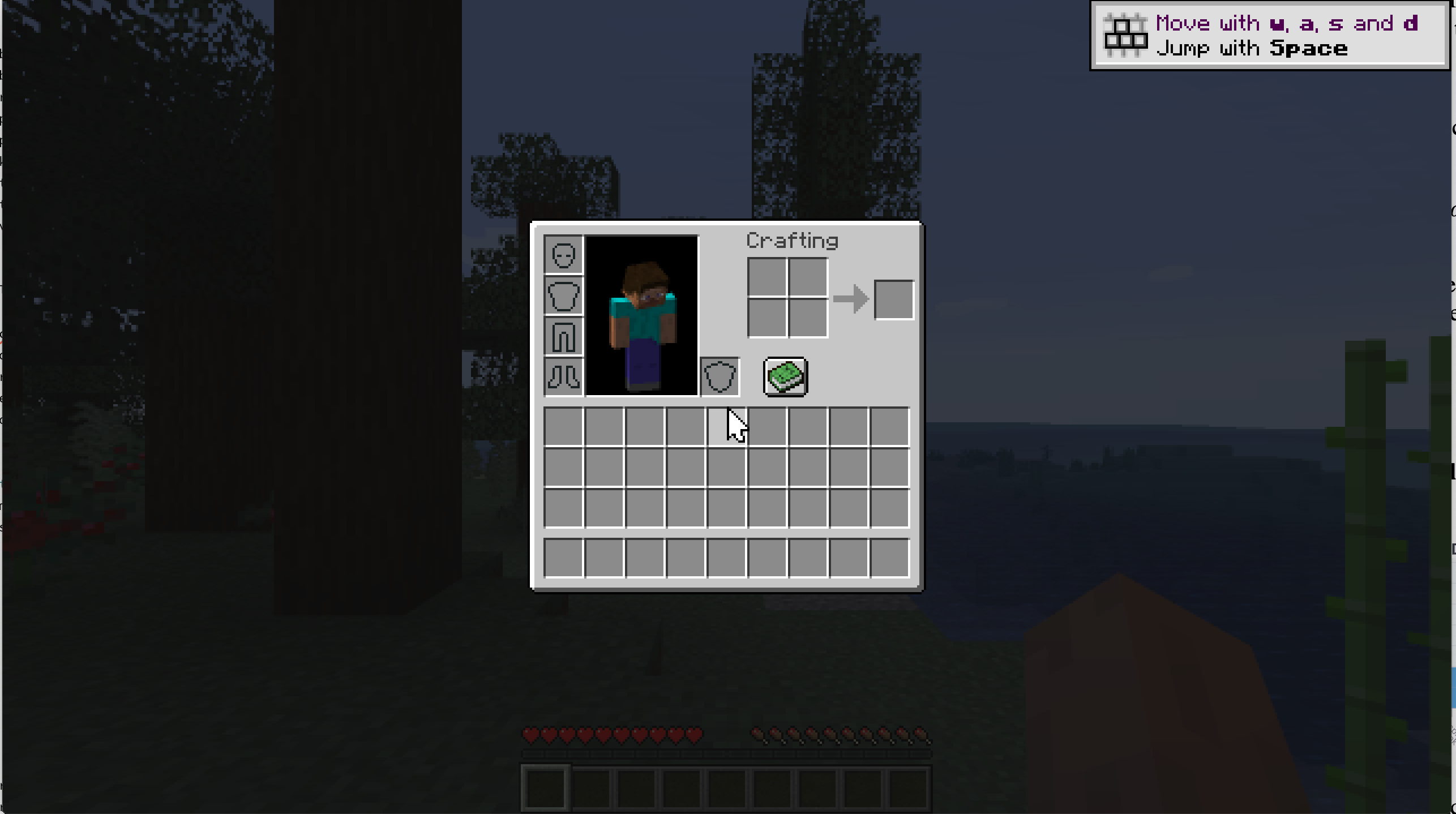} 
    \end{subfigure}
    \begin{subfigure}{0.2\textwidth}
        \centering
        \includegraphics[width=\linewidth]{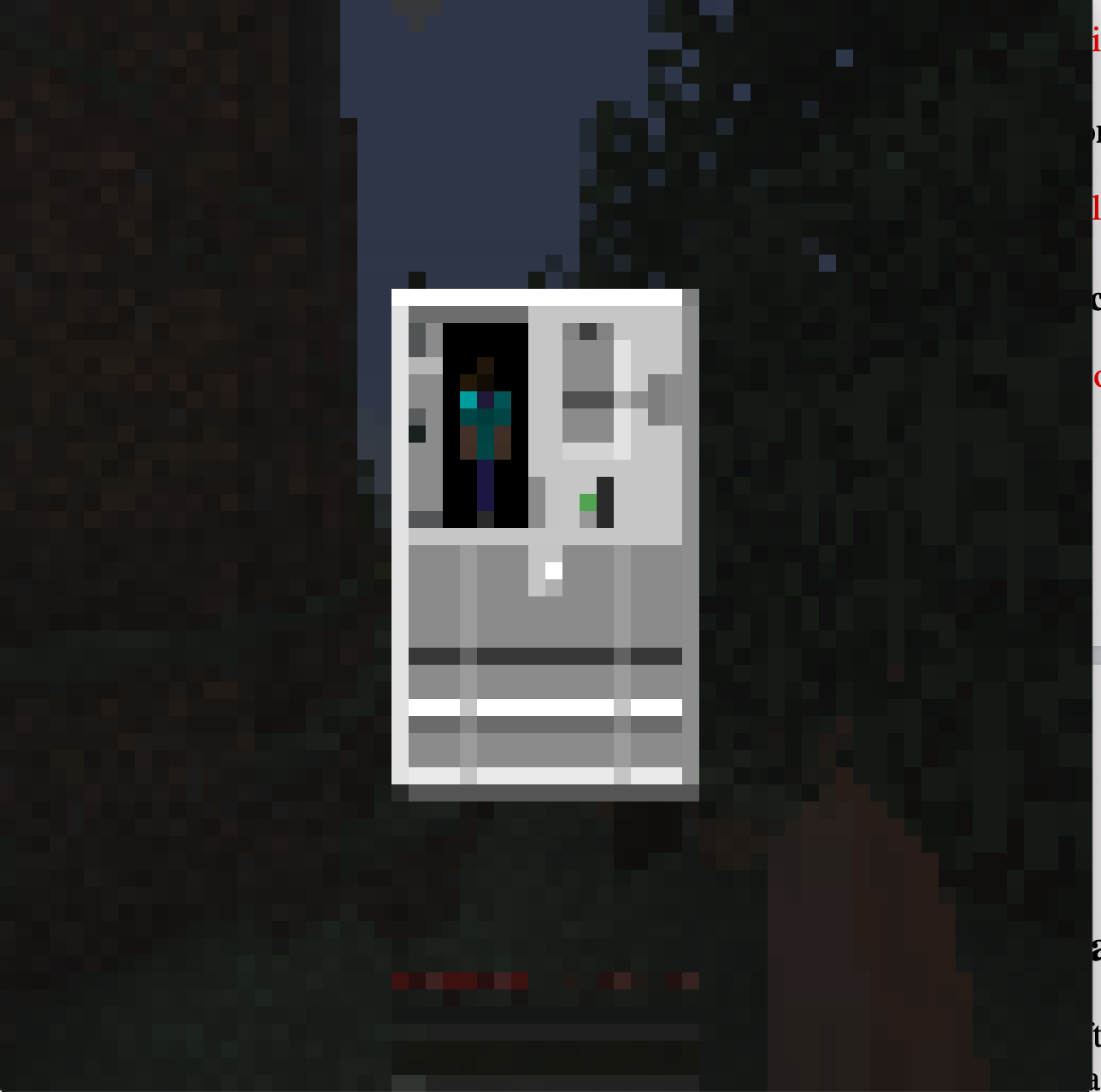} 
    \end{subfigure}
    \begin{subfigure}{0.2\textwidth}
        \centering
        \includegraphics[width=\linewidth]{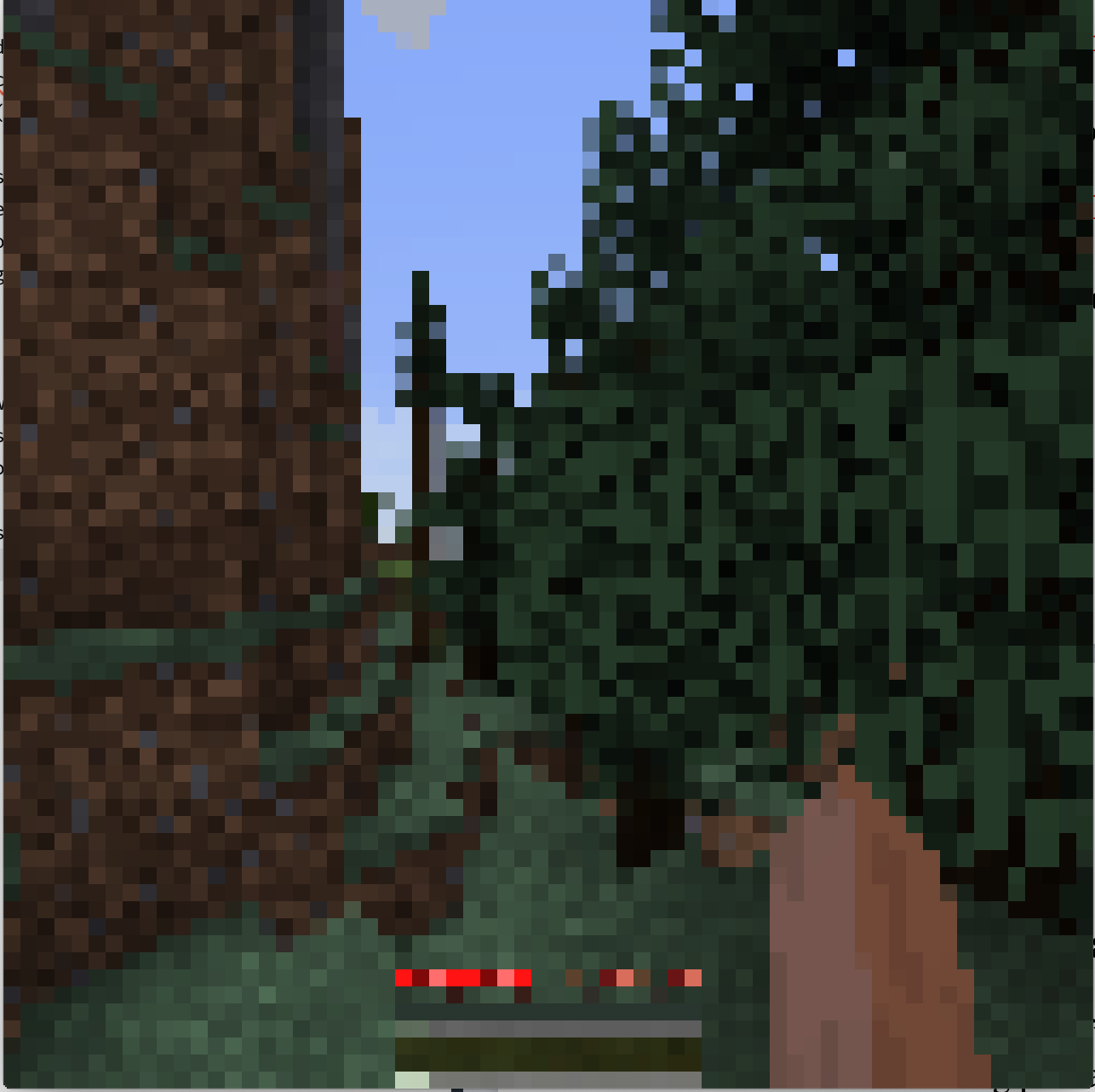} 
    \end{subfigure}
    \caption{\label{fig:a_1_observations}\textbf{(Left)} Sample of a Minecraft frame in the original resolution (640x360) with an in-game GUI open. The mouse cursor can be seen in the center of the image. This particular GUI shows the player's inventory and can be used to craft very basic items. \textbf{(Middle)} We downsample images to 128x128 for computational reasons. Shown is a downsampled observation with an in-game GUI for crafting. This is the resolution used by our models. \textbf{(Right)} A 128x128 observation as seen by our models without in-game GUI. The health, hunger, hotbar overlays, and agent hand can be seen in the lower part of the image.}
\end{figure}
\section{Minecraft environment details}
\label{appendix:minecraft_environment}

Our Minecraft training environment is a hybrid between MineRL \cite{guss2019minerl} and the MCP-Reborn (\href{https://github.com/Hexeption/MCP-Reborn}{github.com/Hexeption/MCP-Reborn}) Minecraft modding package. Unlike the regular Minecraft game, in which the server (or the "world") always runs at 20Hz and the client runs as fast as rendering can complete (typically at 60-100Hz), in our version the client and server run in the same thread at the same frequency. This allows us to run the environment slower or faster than real time, while avoiding artifacts like missing chunks of the world. The action and observation spaces are similar to those of MineRL environments and are described in more detail in the following subsections. The environment also returns diagnostic information, such as in-game stats, contents of the agent's inventory, whether any in-game GUI is open, etc., which we use for tracking and recording but not as inputs to the models. The episode length is 10 minutes for RL experiments and 60 minutes for BC model evaluations. The agent can "die" in a number of ways, such as staying under water for too long and drowning, being killed by hostile mobs, or falling from a tall structure. We do not terminate the episode on agent "death". Instead, just as for humans in the regular Minecraft game, the agent drops all its items when it dies and respawns at a random spot close to the initial spawning spot in the same Minecraft world. The policy state is not masked on death, so the model can remember the fact that it has died and act accordingly.

\subsection{Observation space}
\label{appendix:minecraft_env_observation_space}
The  environment observations are simply the raw pixels from the Minecraft game that a human would see. Unlike MineRL, we do not remove overlays like the hotbar, health indicators, and the animation of a moving hand shown in response to the attack or ``use'' actions. The field of view is 70 degrees, which corresponds to the Minecraft default. GUI scale (a parameter controlling the size of the in-game GUI) is set to 2, and brightness is set to 2 (which is not a Minecraft default, but is very frequently used in online videos).  The rendering resolution is 640x360, which is downsampled to 128x128 before being input to the models. We empirically found 128x128 to be the smallest resolution for which in-game GUI elements are still discernible, and then chose that to minimize compute costs. Whenever an in-game GUI is open, we additionally render an image of a mouse cursor at the appropriate mouse position to match what a human player's operating system does (Fig. \ref{fig:a_1_observations}).

\subsection{Action space}
\label{appendix:action_space}
Our action space includes almost all actions directly available to human players, such as keypresses, mouse movements, and clicks. The specific binary actions we include are shown in Table~\ref{table:binary_actions}.

\begin{table}[htpb]
\begin{center}
\renewcommand\arraystretch{1.4}
\begin{tabular}{ |c|c|p{0.55\linewidth}| } 
\hline
 \textbf{Action} & \textbf{Human action} & \textbf{Description}\\ 
 \hline
 \texttt{forward} & W key & Move forward. \\
 
 \texttt{back} & S key & Move backward.\\ 
 
 \texttt{left} & A key & Strafe left. \\ 
 
 \texttt{right} & D key & Strafe right.\\
 
 \texttt{jump} & space key & Jump.\\
 
 \texttt{inventory} & E key & Open or close inventory and the 2x2 crafting grid. \\
 
 \texttt{sneak} & shift key & Move carefully in current direction of motion. In the GUI it acts as a modifier key: when used with \texttt{attack} it moves item from/to the inventory to/from the hotbar, and when used with \texttt{craft} it crafts the maximum number of items possible instead of just 1.\\
 
 \texttt{sprint} & ctrl key & Move fast in the current direction of motion. \\
 
 \texttt{attack} & left mouse button & Attack; In GUI, pick up the stack of items or place the stack of items in a GUI cell; when used as a double click (attack - no attack - attack sequence), collect all items of the same kind present in inventory as a single stack.\\
 
 \texttt{use} & right mouse button & Place the item currently held or use the block the player is looking at. In GUI, pick up the stack of items or place a single item from a stack held by mouse. \\
 
 \texttt{drop} & Q key & Drop a single item from the stack of items the player is currently holding. If the player presses ctrl-Q then it drops the entire stack. In the GUI, the same thing happens except to the item the mouse is hovering over. \\
 
 \texttt{hotbar.[1-9]} & keys 1 -- 9 & Switch active item to the one in a given hotbar cell. \\
\hline
\end{tabular}
\end{center}
\caption{\label{table:binary_actions}Binary actions included in the action space. \url{https://minecraft.fandom.com/wiki/Controls} has more detailed descriptions of each action.}
\end{table}

One difference between the human action space and our agent's is that we disallow typing arbitrary letters, which is only useful for entering text into the search bar of the crafting recipe book. Humans can either do that or browse the recipe book with the mouse, the latter of which our agent can still do. However, because we do allow the agent to press letters that are also shortcuts for actions (e.g.\ outside of the GUI, the "W" key triggers the \texttt{forward} action) agents are able to press a few keys within the GUI (W, A, S, D, E, Q) that produce letters if the recipe book search bar is selected. We have not seen agents attempt to search the recipe book with these letters. Instead, our agents navigate the recipe book with the mouse or craft by dragging items around the crafting window. 

In addition to the binary (on/off) keypress actions, our action space also includes mouse movements. As with human gameplay, when in-game GUIs are not open, mouse X and Y actions change the agent's yaw and pitch, respectively. When a GUI is open, camera actions move the mouse cursor. Mouse movements are relative (i.e. they move the mouse or camera relative to the current position, and thus their effect depends on the current position).

\begin{figure}[htbp]
\begin{center}
\includegraphics[width=0.5\linewidth]{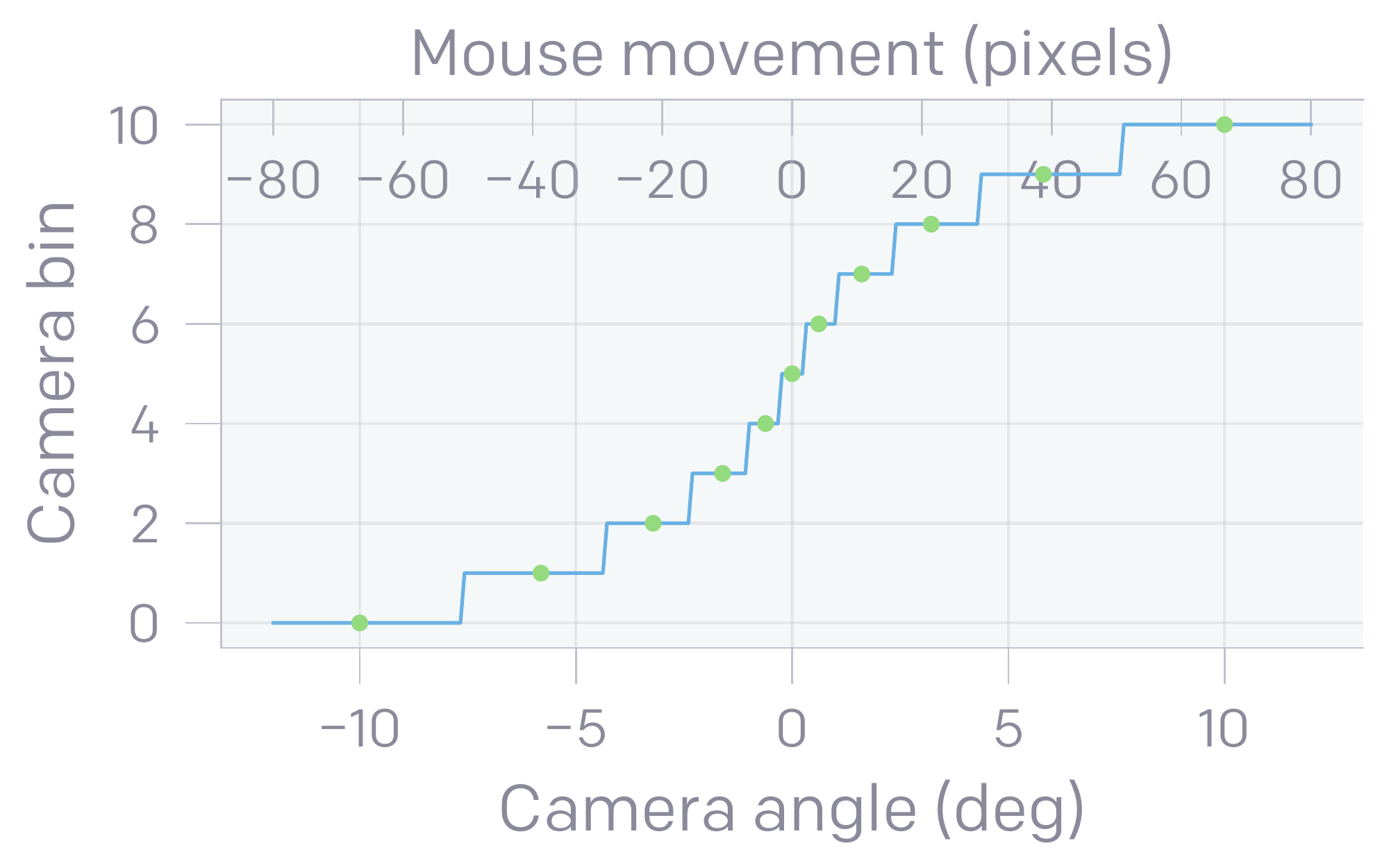}
\end{center}
\vspace{-8px}
\caption{\label{fig:a_1_camera_binning} Relative camera angle or mouse movement in pixels vs.\ action bin. The same binning is used for both X and Y coordinates. The binning is foveated, meaning that binning is more fine-grained for smaller movements and more coarse-grained for larger movements. There are 11 bins for each axis (X and Y). The center of each bin (indicated with green circles) is used when un-discretizing movements (that is, when converting from an action expressed as a bin to a camera angle or mouse movement).}
\end{figure}

Inventory interaction in Minecraft requires fine-grained mouse movements to achieve tasks such as crafting and smelting, while mining and navigating the world can be achieved with coarser mouse action. To be able to achieve both with the same action space, we implemented mouse movements as a set of discrete actions with foveated binning along each axis (Fig. \ref{fig:a_1_camera_binning}), which in preliminary experiments we found to improve crafting performance.

\section{Inverse Dynamics Model Training Details}
\label{appendix:training_details_IDM}
\subsection{IDM Architecture}
\label{appendix:training_details_IDM_architecture}
\begin{figure}[bthp]
\begin{center}
\includegraphics[width=\linewidth]{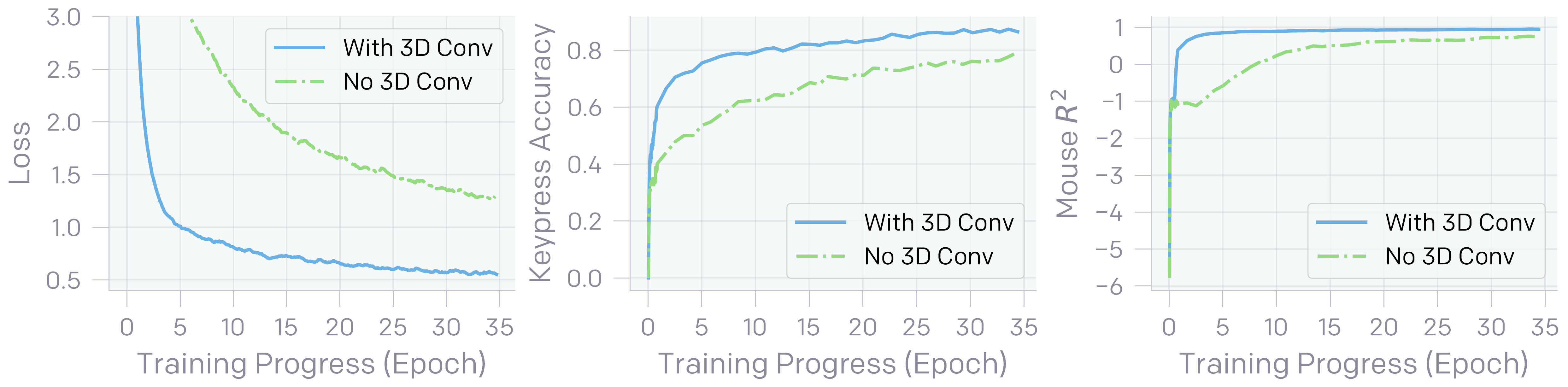}
\end{center}
\caption{Effect of 3-D Convolution in the IDM Architecture.}
\label{fig:idm_3dconv}
\end{figure}

The IDM model has approximately 0.5 billion trainable weights. The input to the IDM is 128 consecutive image frames (128 frames of video), each of which has dimensions $128 \times 128 \times 3$. The IDM is tasked with predicting the action at each frame.
All image pixel values are first divided by 255.0 such that they lie within the range $[0, 1]$. The first layer of the IDM is a 3-D convolution with 128 learnable filters with a temporal kernel width of 5 and spatial kernel widths of 1. This convolution is non-causal, meaning that embeddings at time index $t$ are functions of pixel values at times $t-2$, $t-1$, $t$, $t+1$, and $t+2$. We found this layer to be extremely important in IDM training as it incorporates neighboring temporal information immediately, and we show results comparing IDM performance with and without this layer in Figure~\ref{fig:idm_3dconv}. This comparison was made on the default (1962-hour) IDM dataset.

This initial temporal convolutional layer is followed by a ResNet\cite{he2016deep} image processing network. In this part of the model, no extra temporal information is shared between neighboring frames; however, since each frame was first processed with the temporal convolution, some temporal information is present at this stage. The ResNet image processing network is comprised of three subsequent stacks with widths $W = \{64, 128, 128\}$. Each stack is comprised of, in order, (1) an initial 3x3 convolutional layer with 1-pixel zero padding at the embedding boundary (such that the outgoing embedding dimensions are the same as the incoming embedding dimension) with $W$ output channels, (2) a 3x3 max pooling with stride 2 and padding 1 such that the embedding width and height are halved, and (3) two classic ResNet blocks as defined in \citet{he2016deep} with each layer also having $W$ output channels.

The output of the ResNet stack is flattened into a 1-dimensional vector of size $2^{17} = 131072$ (one vector for each frame in the video) such that at this stage there are 128 vectors of size 131072. Each vector is independently processed with two frame-wise dense layers with 256 output activations and then 4096 output activations, respectively. The result is then fed through 4 subsequent non-causal (umasked) residual transformer\cite{vaswani2017attention} blocks. Each block first has an unmasked attention layer, i.e. frames may attend to future frames, with 32 attention heads of dimension 128 each and a surrounding residual connection that skips this layer. The embedding is then passed through a frame-wise dense layer with output dimension 16384 and another with output dimension returning to 4096; a single residual connection skips past this pair of frame-wise dense layers (not skipping past each layer separately, but skipping the pair). All dense layers have their weights tied through time, so each frame in the video is processed with the same weights.

Finally, independent dense layer heads for each action are pulled from the final embedding -- a 2 class on/off categorical parameterized with a softmax for each available key as well as a 11-way categorical for both the discretized horizontal and vertical mouse movements (See Appendix~\ref{appendix:action_space} for details on the action space).

Each dense layer or convolutional layer in the network is preceded by a layernorm\cite{ba2016layer} and followed by a ReLU non-linearity. Weights are initialized with Fan-In initialization\cite{lecun2012efficient} and biases are initialized to zero. 

\subsection{IDM Training}
\label{appendix:training_details_IDM_training}
The total loss for the network is the sum of each independent action prediction loss (one for each key and one for both mouse directions). Each independent loss is the negative log-likelihood of the correct action. We use the ADAM\cite{kingma2014adam} optimizer with a linear learning rate decay.
We use an initial learning rate of 0.003, a batch size of 128 (where each item in the batch is a video sequence of 128 frames), and a weight decay of 0.01. 
Hyperparameters were tuned in preliminary experiments. The IDM is trained on our contractor collected dataset for 20 epochs. This took 4 days on 32 A100 GPUs.

We add data augmentation to each video segment; augmentations are randomly sampled once per segment such they are temporally consistent. Using the Pytorch\cite{pytorch} transforms library, we adjust the hue by a random factor between -0.2 and 0.2, saturation between 0.8 and 1.2, brightness between 0.8 and 1.2, and contrast between 0.8 and 1.2. We also randomly rotate the image between -2 and 2 degrees, scale it by a random factor between 0.98 and 1.02, shear it between -2 and 2 degrees, and translate it between -2 and 2 pixels in both the $x$ and $y$ dimensions.

Due the large computational cost of running all of the experiments in this paper, training results are from one run of training (for IDM, BC, and RL training): this non-ideal situation is mitigated because deep learning training tends to be low variance \cite{dauphin2014identifying, yosinski2014transferable} and because we often have data points from sweeps (e.g.\ on dataset size) that suggest overall trends.

\subsection{Generating Pseudo Labels with the IDM}
Section~\ref{section:results_IDM} shows that inverse dynamics modeling is a much easier task than behavioral cloning because IDMs can be non-causal. The IDM is trained to simultaneously predict all 128 actions for each video sequence, so the IDM will effectively be causal for frames at the end of the video clip because future frames are not included in the sequence. For this reason, we apply the IDM over a video using a sliding window with stride 64 frames and only use the pseudo-label prediction for frames 32 to 96 (the center 64 frames). By doing this, the IDM prediction at the boundary of the video clip is never used except for the first and last frames of a full video.

\section{Foundation Model Behavioral Cloning}
\subsection{Foundation Model Architecture}
The behavioral cloning model architecture is the same as the IDM architecture described in Appendix \ref{appendix:training_details_IDM_architecture} except that we modify the architecture so that it is causal (i.e. cannot see the future when making predictions). This means the BC architecture does not have the initial non-causal convolution the IDM has (this layer is omitted completely). Furthermore, the residual transformer layers are now causally masked (as is standard in language modeling) and we do Transformer-XL-style\cite{dai2019transformer} training where frames can attend to keys and values from past batches within the same video. We also use a Transformer-XL-style relative attention position embedding.

\subsection{Null Action Filtering}
The most common action humans take is the null action (no keypresses or mouse movements), which accounts for 35\% of all actions they take. Among other reasons, a player may take the null action to wait for something in the game to finish, to pause between actions, or to take a break to grab a glass of water. Early on in the project we found that the BC model would take a much larger fraction than 35\% of null actions, often upwards of 95\%. In order to prevent this behavior we removed frames with null actions from the dataset. We compare a few different treatments: we filter nulls if there have been 1, 3, or 21 frames of consecutive null actions, and include a treatment that does not perform any null filtering. Null action filtering generally helps, increasing all crafting rates (Figure~\ref{fig:a_null_and_joint} left).
Filtering only groups of 3 performed slightly better than filtering all null action or groups of 21. Initial experiments indicated that filtering all null actions was better; however, after further model tuning and after we had already trained our largest models, we found that filtering only groups of 3 or more null actions performed best. Due to compute constraints we were not able to redo all experiments with this setting, but doing so would be a reasonable choice for any future work.

\subsection{Joint Hierarchical Action Space}
We originally worked with a factored action space, where each keypress could be independently on or off, and this choice was independent of whether the mouse was being moved. This could cause issues for modeling the human behavior distribution exactly. Say for a given state, humans either with 50\% probability (a) move forward and attack or with 50\% probability (b) move left and drop their item. The best a factored distribution can do is to assign 50\% probability to each of the 4 constituent actions because it chooses to press each button simultaneously and independently. See Appendix~\ref{appendix:action_space} for details on the entire action space.

\begin{figure}[t]
\begin{subfigure}{0.5\textwidth}
    \centering
    \includegraphics[width=\linewidth]{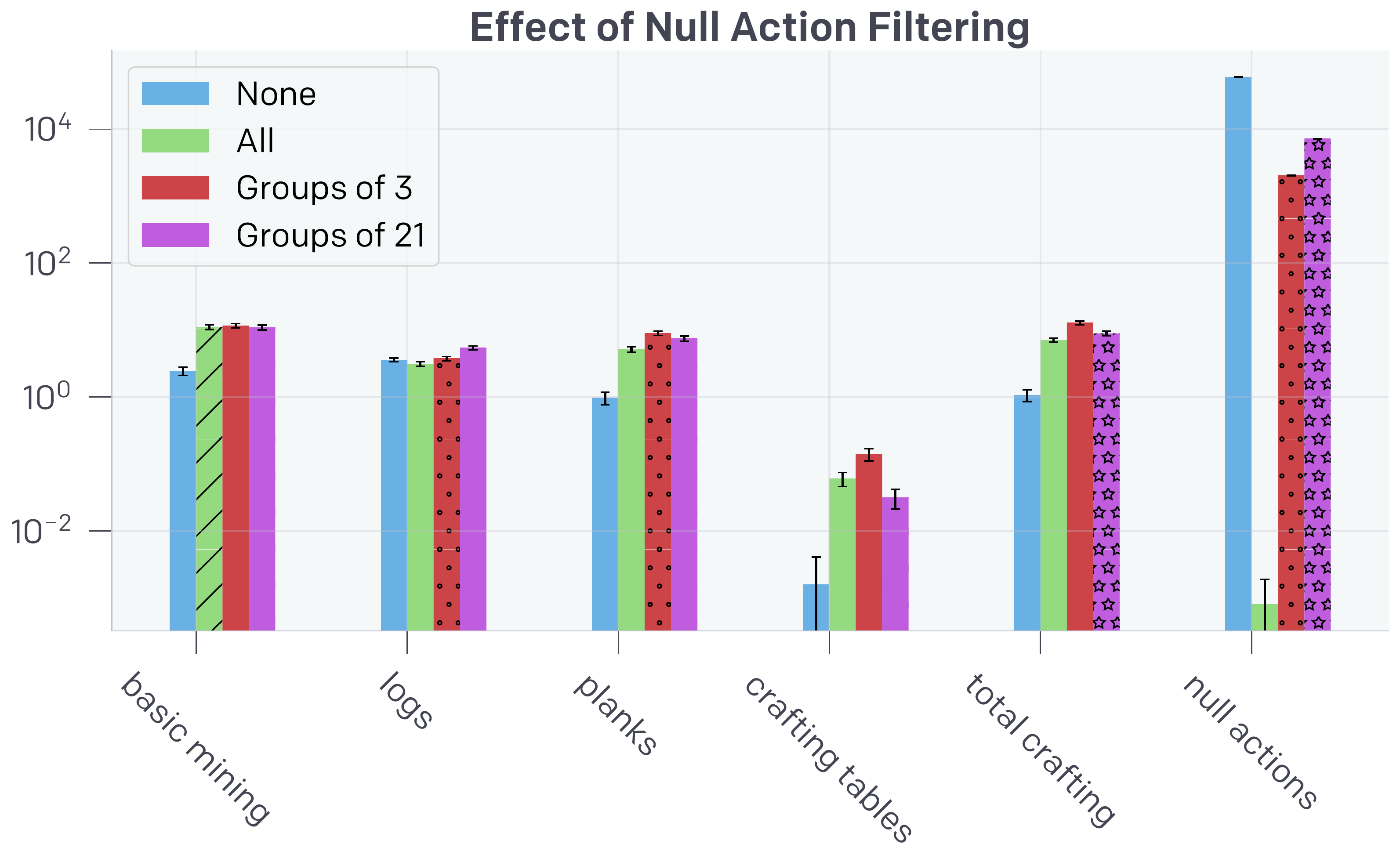} 
\end{subfigure}
\begin{subfigure}{0.5\textwidth}
    \centering
    \includegraphics[width=\linewidth]{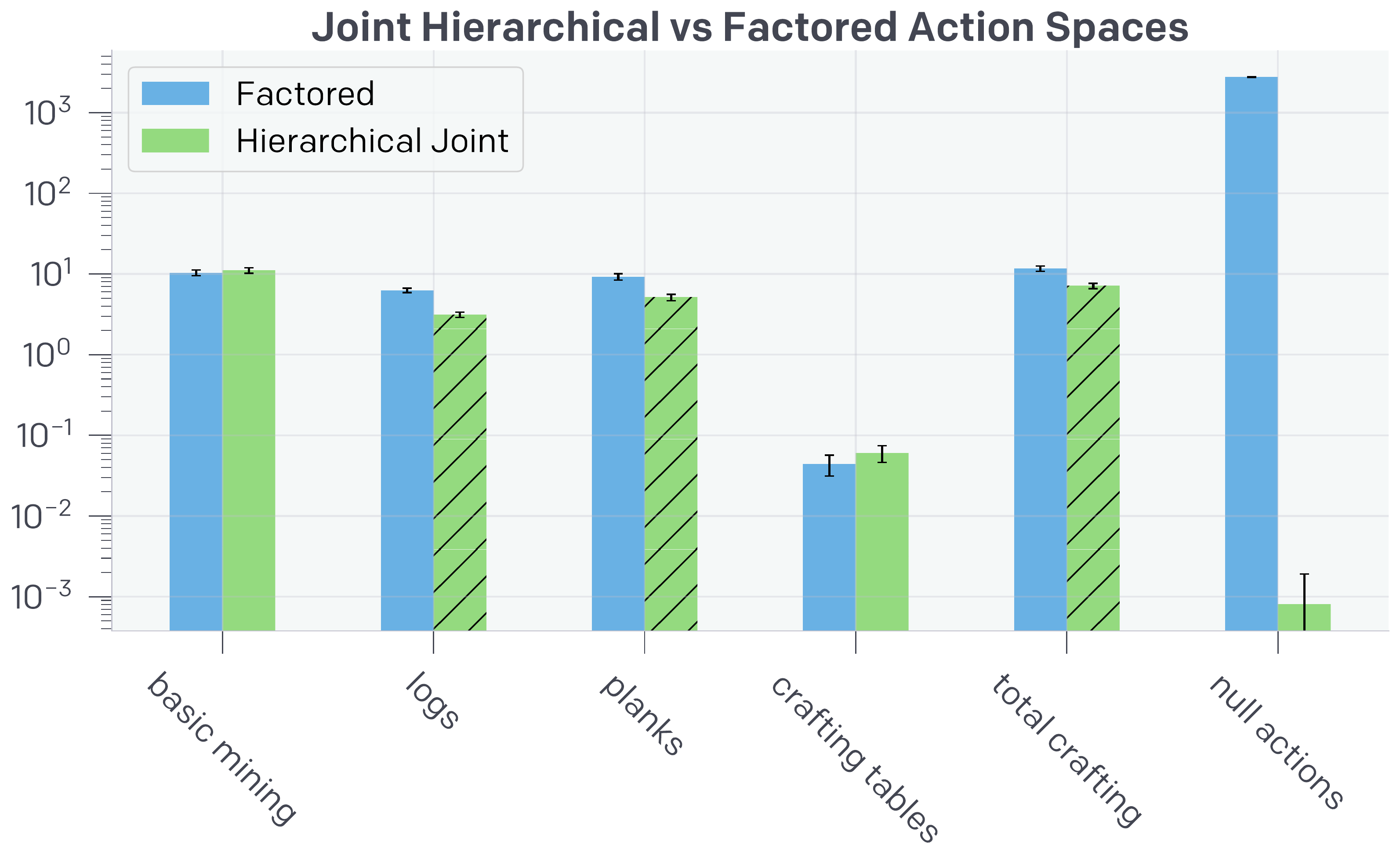} 
\end{subfigure}
\caption{\textbf{(Left)} Effect of Null Action Filtering during training. We compare environment metrics and number of sampled null action during rollouts (rightmost group of columns) for the following treatments: no null action filtering (blue), filtering all null actions (green), filtering only groups of 3 or more null actions (red), and filtering only groups of 21 or more null actions (purple).
\textbf{(Right)} Hierarchical versus Factored Action Spaces.}
\label{fig:a_null_and_joint}
\end{figure}

For this reason, we implemented a joint distribution over actions; however, the full joint distribution over 20 binary buttons and two mouse movement dimensions discretized into 11 bins each would result in in $2^{20} \times 11^2 \approx 1.2\times10^8$ possible combinations. This is far too large for many reasons, e.g.\ the final layer from  the transformer stack with a dimension of 4096 would need to be mapped to each combination resulting in $4096 \times 1.2\times10^8 \approx 5.2 \times 10^{11}$ parameters for this final layer alone. In order to reduce this we noted that many buttons in Minecraft have no effect when simultaneously pressed; for example, if a player tries to move forward and backward at the same time, they remain in place. Below we list the the sets of mutually exclusive actions. Furthermore, the inventory button is exclusive with all other buttons and mouse movement.

\begin{center}
\begin{tabular}{ |c|c|p{0.55\linewidth}| } 
\hline
 \textbf{Mutually Exclusive Actions}\\ 
 \hline
 \texttt{forward}, \texttt{back} \\
 \texttt{left}, \texttt{right} \\
 \texttt{sprint}, \texttt{sneak} \\
 \texttt{hotbar.[1-9]} \\
\hline
\end{tabular}
\end{center}

Even reducing the joint action space to reflect these mutually exclusive combinations still results in a huge action space when combined with the discretized mouse movements, i.e. $3^3\times10\times2^4\times11^2 + 1 \approx 5.2\times10^5$. 
This calculation results from $3^3$ for the 3 sets of 2 mutually exclusive keys above where taking neither in the set is an option, $\times10$ for the 9 hotbar keys or no hotbar keypress, $\times2^4$ for the remaining binary 4 keys: \texttt{use}, \texttt{drop}, \texttt{attack}, and \texttt{jump}, $\times11^2$ for mouse movements, and finally $+1$ for the \texttt{inventory} button which is mutually exclusive with all other actions. $\sim5.2\times10^5$ is still quite large so we chose to implement a hierarchical binary action for camera being moved or not. If this action is on, then there is a secondary discrete action head with 121 classes (the joint distribution of mouse movements because each discretized mouse direction has 11 bins) that determines where to move the mouse. If the hierarchical action is off, then there is no mouse movement, loss for the secondary mouse movement action is masked during training, and the secondary action head need not be sampled during evaluations. While this no longer models the full joint distribution, 
it is quite a bit better than the factored action space since dependencies between keypresses as well as whether or not to move the mouse (although not which mouse movement) are modeled jointly. The resulting action space has dimension $3^3\times10\times2^4\times 2 + 1 = 8461$ (the $11^2$ dimensional multiplier for camera movement has been replaced by a multiplier of 2 here, corresponding to a binary action for whether or not to move the mouse) with an additional 121-dimension head for the joint camera movements. In the future it would be interesting to implement sequential conditional action spaces to more completely model the joint distribution.

In Figure~\ref{fig:a_null_and_joint} (right) we compare environment rollout performance between BC models with the hierarchical joint action space and with the factored action space. Environment statistics are fairly comparable; however, we see that the factored action space model samples far more null actions. This is an important example of the factored action space failing to correctly model the distribution in the dataset because, due to null action filtering, there are 0 null actions in the dataset these models train on. Despite this, the factored model samples many null actions because the prediction for each key is not conditioned on other keypresses.

\subsection{Foundation Model Training}
The foundation model training is similar to the IDM training, with the exception of labels being IDM-generated pseudo labels. The hyperparameters used for foundation model training are listed in Table \ref{table:a_foundation_hps}.

\begin{table}[h]
\begin{center}
\begin{tabular}{ |c|c|p{0.55\linewidth}| }
\hline
 \textbf{Hyperparameter} & \textbf{Value} \\  
 \hline
 Learning rate & 0.002147 \\
 Weight decay & 0.0625 \\
 Epochs & 30 \\
 Batch size & 880 \\
 \hline
\end{tabular}
\end{center}
\caption{\label{table:a_foundation_hps}Hyperparameters for foundation model training}
\end{table}

\section{Behavioral Cloning Fine-Tuning}
Behavior cloning fine-tuning is similar to the foundation model training, except we either use a focused subset of all the videos (\texttt{early\_game} dataset, described in \ref{appendix:early_game_data}) with pseudo labels, or contractor data (\texttt{contractor\_house} dataset, described in \ref{appendix:contractor_house_data}) with ground-truth labels.
The hyperparameters used for behavior cloning fine-tuning are listed in Table \ref{table:a_bc_hps}. We used 16 A100 GPUs for about 6 hours when fine-tuning on \texttt{contractor\_house} dataset, and 16 A100 GPUs for about 2 days when fine-tuning on \texttt{early\_game} dataset. 

\begin{table}[h]
\begin{center}
\begin{tabular}{ |c|c|p{0.55\linewidth}| }
\hline
 \textbf{Hyperparameter} & \textbf{Value} \\  
 \hline
 Learning rate & 0.000181 \\
 Weight decay & 0.039428 \\
 Epochs & 2 \\
 Batch size & 16 \\
 \hline
\end{tabular}
\end{center}
\caption{\label{table:a_bc_hps}Hyperparameters for behavior cloning fine-tuning}
\end{table}
\section{Reinforcement Learning Fine-Tuning}

\subsection{Reinforcement Learning Fine-Tuning Training Details}
\label{Appendix:RLTrainingDetails}

RL experiments were performed with the phasic policy gradient (PPG) algorithm,\cite{pmlr-v139-cobbe21a} an RL algorithm based on the proximal policy optimization (PPO) algorithm\cite{schulman2017proximal} that increases sample efficiency by performing additional passes over the collected data to optimize the value function as well as an auxiliary value function.
These algorithms have been described extensively in previous work,\cite{pmlr-v139-cobbe21a, schulman2017proximal} so here we describe them only briefly.
A major inefficiency when training on-policy algorithms is that, to remain on-policy, one can only take a single gradient step before new rollout data needs to be gathered to continue optimization.
To alleviate the potentially destructive effects of taking multiple optimization steps in a single iteration, PPO prevents the policy from changing too much in a single step by clipping the loss when the difference between the current policy and the policy before the update becomes too large.\cite{schulman2017proximal}
We also use generalized advantage estimation (GAE), which can speed-up credit assignment by looking more than 1 step into the future when determining the advantage of an action, with the look-ahead being determined by hyperparameter $\lambda$.\cite{schulman2015high}

PPG improves the sample efficiency of PPO when the policy and value function share the same network by following different optimization processes for the policy, the value function, and their shared representation.
PPG splits optimization in two phases: a wake phase and a sleep phase.
In the wake phase, the policy and value function are optimized as in normal PPO training, with the only exception being that every sample is used at most once, which prevents the policy from overfitting on these samples.
In the sleep phase PPG optimizes the value function and an auxiliary value function (which is optimized with the exact same loss as the regular value function, but its output is never used during training), while keeping a Kullback-Leibler (KL) divergence loss to the policy before the start of the sleep phase to ensure that the policy does not change. Because the policy is not optimized in this step, PPG does allow samples to be reused multiple times in this phase.
The assumption behind optimizing the value function during the sleep phase is that value function optimization is less sensitive to being trained multiple times on the same sample.
Optimizing the auxiliary value function does not directly affect either the value function or the policy, but it can improve the shared representation of both functions (the assumption being that predicting the value-function requires encoding all features that are important for distinguishing states).
The coefficients for the three losses (value function loss, auxiliary value function loss, and KL loss) are listed in Table~\ref{tab:rlhyperparameters}.
In our experiments a single iteration consists of two sleep cycles and one wake cycle.

Because the value and auxiliary value functions are not optimized during BC pre-training, they are initialized at the start of RL fine-tuning. 
Each value function is implemented as a single, fully connected layer on top of the last residual transformer block of the pretrained model (Appendix~\ref{appendix:training_details_IDM_architecture}).
The weights of the auxiliary value function are randomly initialized while the weights of the regular value function are initialized with zero weights, which appeared to prevent destructive updates early in training that could happen with a randomly initialized value function.
To prevent the value-function loss from having gradients that depend greatly on the magnitude of the reward, we normalize the value-function target by subtracting the mean and dividing by the standard deviation, which are estimated through an exponentially weighted moving average.

\begin{table}[h]
\begin{center}
\begin{tabular}{|l|l|}
\hline
 \textbf{Hyperparameter} & \textbf{Value} \\  
 \hline
Learning rate:           & $2\times 10^{-5}$ \\
Weight decay:            & $0.04$     \\
Batch size:              & $40$       \\
Batches per iteration:   & $48$       \\
Context length:          & $128$      \\
Discount factor ($\gamma$): & $0.999$    \\
GAE $\lambda$:            & $0.95$     \\
PPO clip:                & $0.2$      \\
Max Grad norm:           & $5$        \\
Max Staleness:           & $2$        \\
PPG sleep cycles:        & $2$        \\
PPG sleep value-function coefficient:        & $0.5$        \\
PPG sleep auxiliary value-function coefficient:        & $0.5$        \\
PPG sleep KL coefficient:        & $1.0$        \\
PPG sleep max Sample Reuse:    & $6$        \\
KL divergence coefficient $\rho$:                 & $0.2$      \\
Coefficient $\rho$ decay:                & $0.9995$   \\  
\hline
\end{tabular}
\end{center}
\caption{Hyperparameters for RL experiments. These are the hyperparameters for all treatments with two exceptions. First, when fine-tuning from the early-game model without a KL divergence loss, in addition to the KL divergence loss being set to 0, the learning rate was set to $3\times 10^{-6}$ (the best setting out of a sweep over 5 different learning rates), as we found that performance was substantially lower with the standard learning rate of $2\times 10^{-5}$ and the agent did not even learn to collect logs. We suspect that the reason that the learning rate needed to be lowered when fine-tuning without a KL loss is that the KL loss prevents making optimization steps that change the policy too much in a single step, especially in early iterations when the value function has not been optimized yet, and the KL loss thus makes it possible to optimize with a higher learning rate. Second, when running RL from a randomly initialized policy there is no KL divergence loss or KL divergence decay, but instead we use an entropy bonus of $0.01$, which reportedly worked well in previous work.\cite{kanitscheider2021multi}}
\label{tab:rlhyperparameters}
\end{table}

To prevent catastrophically forgetting the skills of the pretrained network when RL fine-tuning, we apply an auxiliary KL divergence loss between the RL model and the frozen pretrained policy.\cite{vinyals2019grandmaster}
This loss is defined as:
\begin{equation}
    L_{klpt} = \rho \mathrm{KL}(\pi_{pt}, \pi_\theta)
\end{equation}
Where $\pi_\theta$ is the the policy being trained, $\pi_{pt}$ is the frozen pretrained policy, $\mathrm{KL}(\pi_{pt}, \pi_\theta)$ is the Kullback-Leibler divergence between the policy being trained and the pretrained policy, and $\rho$ is a coefficient to weight this loss relative to other losses.

In the fine-tuning experiments, this KL divergence loss replaces the common entropy maximization loss, which is often added to RL experiments to encourage exploration.\cite{williams1992simple, mnih2016asynchronous}
The idea behind entropy maximization is that, when all actions appear to have equal value, such as when the agent has not learned about the next reward, it should maximize its entropy to increase the chance that it discovers the next reward.
Blindly exploring by maximizing entropy is effective when the state and action spaces are sufficiently small or the reward is sufficiently dense, but becomes infeasible when the state and action spaces are large and rewards are sparse, which is the case in the diamond-pickaxe task.
Instead of blindly exploring through uniform-random actions, we assume that the pretrained policy has an action distribution that is much more likely to take sequences of actions that lead to interestingly new states, and thus, in states where the agent assigns equal value to each of its actions, it should mimic the action-distribution of the pretrained policy instead of a uniform-random action distribution.
In experiments with a randomly initialized policy we do include the entropy maximization loss with a coefficient of $0.01$, which has been an effective setting in other Minecraft work.\cite{kanitscheider2021multi}
Empirically, we found that a high coefficient $\rho$ for this KL divergence loss would prevent the agent from properly optimizing the reward function while a low coefficient $\rho$ was ineffective at protecting the learned skills of the pretrained policy and preventing catastrophic forgetting.
As such, we start with a relatively high coefficient $\rho$ and decay it % \pz{"loss" here is actuall a loss coefficient, correct? Such that $loss = KL\_loss * KL$ ? I think the phrase "decay the loss" is a bit confusing, given that it is $KL\_loss$ coefficient that is decayed. I understand that calling it "KL coefficient" does not work because PPG/PPO has its own KL coefficient... Maybe a greek letter, and add an equation? This will also clarify the order of policies in KL (as in, is it KL(current || pretrained) or KL(pretrained || current)? }\ale{Agreed}
by a fixed factor after each iteration (Table~\ref{tab:rlhyperparameters}).
This method protects policy skills in early iterations while guaranteeing that the policy can eventually maximize the reward function, regardless of how different its behavior has to be to do so relative to the pretrained policy.

For the reward function we estimated the rough quantities of each item that a human player might gather when trying to craft a diamond pickaxe, and we reward the model for gathering up to that quantity for each item.
We started these estimates by iterating over the technology tree backward from a diamond pickaxe and adding the requirements for each item to the reward function (e.g.\ first we added a diamond pickaxe to the reward function, then we added the 3 diamonds and 2 sticks required for crafting a diamond pickaxe, then we added the 1 iron pickaxe required for mining diamonds, and so on).
Then we added coal and torches to the reward function, with coal being useful as fuel when smelting iron and for crafting torches while the torches themselves improve visibility and prevent enemies from spawning. 
Finally, we reward the model for bringing additional logs (5 logs are required to craft all items in the reward function, but we reward up to 8 logs), which can be used as fuel or crafted into a crafting table or sticks if the agent runs out.
In practice the agent rarely collects the additional logs, places the torches, or uses coal as fuel when smelting, but the reward function was based on human expectations on what would be useful to execute this task, rather than designed around how an RL model behaves after training.
Finally, to encourage the agent to keep mining diamonds and crafting diamond pickaxes after it has crafted its first diamond pickaxe, we did not put a limit on the number of diamonds or diamond pickaxes that would be rewarded.

\begin{table}[]
\begin{center}
\begin{tabular}{|l|r|r|}
\hline
\textbf{Item} & \textbf{Quantity rewarded} & \textbf{Reward per item} \\
\hline
Log & $8$ & $1/8$ \\
Planks & $20$ & $1/20$ \\
Stick & $16$ & $1/16$ \\
Crafting table & $1$ & $1$ \\
Wooden pickaxe & $1$ & $1$ \\
Cobblestone & $11$ & $1/11$ \\
Stone pickaxe & $1$ & $1$ \\
Furnace & $1$ & $1$ \\
Coal & $5$ & $2/5$ \\
Torch & $16$ & $1/8$ \\
Iron ore & $3$ & $4/3$ \\
Iron ingot & $3$ & $4/3$ \\
Iron pickaxe & $1$ & $4$ \\
Diamond & inf & $8/3$ \\
Diamond pickaxe & inf & $8$ \\
\hline
\end{tabular}
\end{center}
\caption{Reward per item and total quantity rewarded.}
\label{tab:rlreward}
\end{table}

The rewards for the different items are separated into 4 tiers, roughly depending on how late a player would usually get the relevant item.
The first tier consists of all wooden and stone items and has a base reward of 1, the second tier consists of all items requiring coal with a base reward of 2, the third tier consists of all items requiring iron with a base reward of 4, and the final tier is diamond with a base reward of 8.
Thus items later in the sequence of items towards a diamond pickaxe generally give a higher reward.
To make sure that the agent does not over-value items that are supposed to be gathered in bulk (e.g.\ the agent is rewarded for up to 20 planks but only up to 1 crafting table, which can cause the agent to focus on planks at the expense of creating a crafting table), we divide the base reward of each item by the total quantity that the agent gets rewarded for (for the purpose of determining the reward, the total quantity for diamonds is 3 and the total quantity for diamond pickaxes is 1, even though we did not put a limit on the number of these items being rewarded). For example, the agent is rewarded for 3 iron ore, which has a base reward of 4 for being in the iron tier and up to 3 blocks of iron ore are rewarded, thus the reward per block of iron ore is $4/3$.
The quantity and reward for each item are listed in Table~\ref{tab:rlreward}.

%\jh{The next paragraph can be cut as well, given that it was added specifically to make it easier to explain the crafting-table bug, but it does stand on its own, so I am also fine with keeping it.}\bb{may as well leave it}\jc{agreed, let's leave it}
While every item in the sequence towards a diamond pickaxe is rewarded, the reward function is still sparse and, in some cases, even deceptive. The sparsity comes from the fact that it can take thousands of actions to find the next reward, even after the agent has acquired all the necessary prerequisites (e.g.\ human players often take more than 10,000 actions to find a diamond after crafting an iron pickaxe).
The reward function can be deceptive when the most efficient method for getting one item can make it far more difficult to get the next item.
For example, a good strategy for the agent to craft a stone pickaxe quickly is to mine (i.e. spend a few seconds to pick up) its crafting table after crafting a wooden pickaxe, such that the agent has immediate access to a crafting table as soon as it has collected enough cobblestone.
However, the fastest way to get a reward for gathering cobblestone is to mine down immediately after crafting a wooden pickaxe, while leaving the crafting table behind.
Thus following the optimal strategy for gathering cobblestone makes it more difficult to learn to craft a stone pickaxe.

Experiments ran for approximately 6 days (144 hours) on 80 GPUs (for policy optimization) and 56,719 CPUs (mostly for collecting rollouts from Minecraft). In this time the algorithm performed roughly 4,000 optimization iterations and collected roughly 1.4 million Minecraft episodes consisting of 12,000 frames each, for a total of 16.8 billion frames.

\subsection{Reinforcement Learning Fine-Tuning Additional Data}
\label{Appendix:RLPreliminary}

\begin{figure}[h]
\begin{center}
\includegraphics[width=0.5\linewidth]{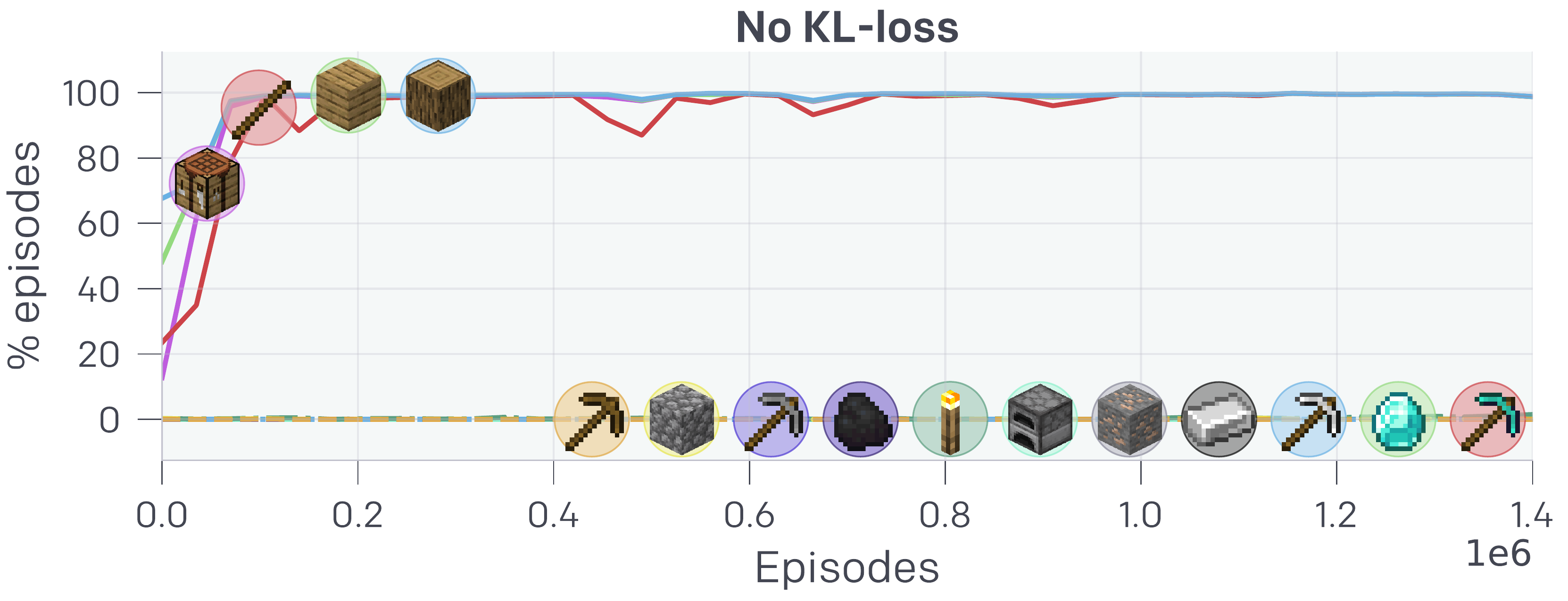}
\end{center}
\caption{Items obtained when RL fine-tuning from the early-game model without a KL loss. The model learns to obtain all items that the early-game model can craft zero-shot, which are logs, planks, sticks, and a crafting table. In contrast to the treatment with a KL-penalty, it does not learn any items beyond these initial four, likely because skills that are not performed zero-shot, and for which the model thus does not initially see any reward, are catastrophically forgotten while the first four items are learned.}
\label{fig:rlft_nokl}
\end{figure}

\begin{figure}[h]
\begin{center}
\includegraphics[width=0.49\linewidth]{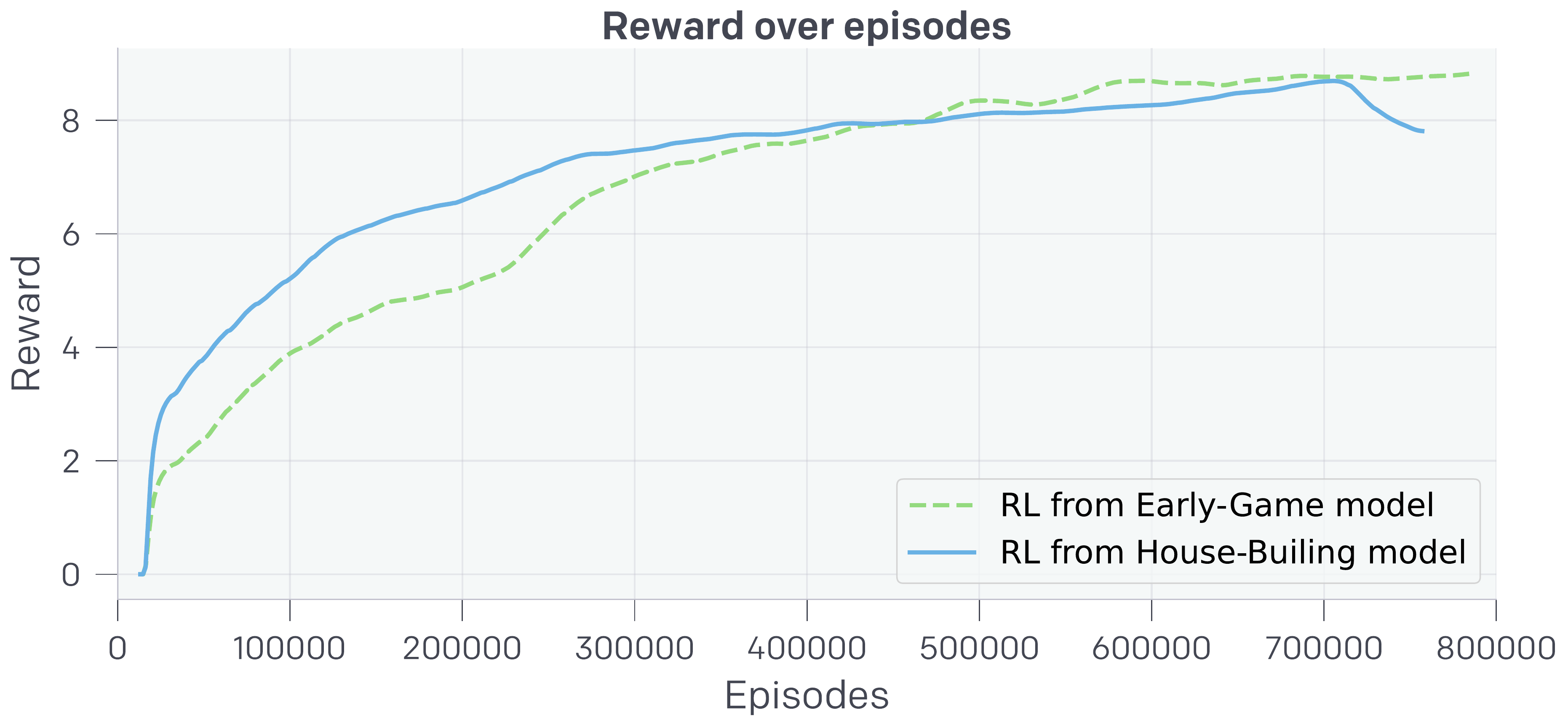}
\includegraphics[width=0.49\linewidth]{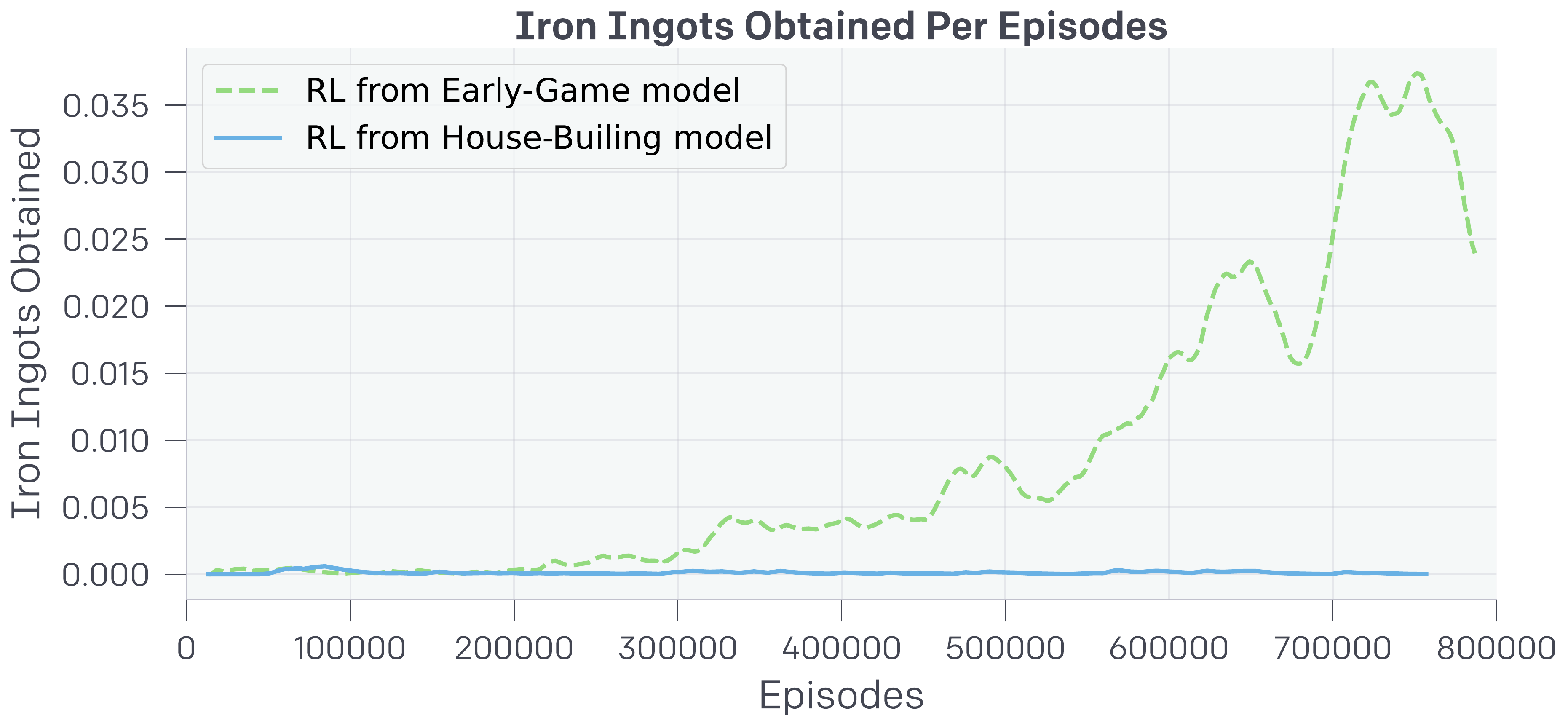}
\end{center}
\caption{Preliminary experiments when RL fine-tuning from the early-game model compared to RL fine-tuning from the house-building model. \textbf{(Left)} While reward initially increases faster when fine-tuning from the house-building model, fine-tuning form the early-game model eventually obtains a slightly higher reward. \textbf{(Right)} RL fine-tuning from the early-game model has a higher likelihood of smelting an iron-ingot, which is why the early-game model was chosen for future RL fine-tuning experiments. }
\label{fig:rlft_vs_housebuilding}
\end{figure}

Additional figures that are helpful for understanding the main results of the RL fine-tuning experiments are presented in this section.
First, we show the items-over-training figure when RL fine-tuning from the early-game model without a KL loss (Fig.~\ref{fig:rlft_nokl}).
When training without a KL loss, the model only learns to obtain the four items that the early-game model is capable of getting zero-shot, which are logs, planks, sticks, and crafting tables.
Second, we present preliminary experiments in which we directly compare RL fine-tuning from the house-building model and RL fine-tuning from the early-game model (Fig.~\ref{fig:rlft_vs_housebuilding}).
These experiments differ from the main experiments in that, for both treatments shown here, the KL loss coefficient was set to $0.4$, the learning rate was set to $6\times 10^{-5}$, and the reward for each item was $1/quantity$ for all items (i.e. items closer to the diamond pickaxe did not have an increased reward).
While RL fine-tuning from the house-building model initially worked better than RL fine-tuning from the early-game model, fine-tuning from the early-game model worked better after 800,000 episodes and showed signs of smelting iron ingots, which is why the early-game model was chosen for the main experiments.

\section{Foundation Model Scaling}
\label{Appendix:FoundationModelScaling}

In early experiments we found that increasing model size led to models staying in the efficient learning regime longer into training.\cite{kaplan2020scaling} Here we compare the 0.5B model described in Section~\ref{section:results_foundation} to both a 248M and 71M parameter model. Both of these models are trained for 15 epochs as compared to the 30 epochs the 0.5B model trained for. These models have the same architecture as the 0.5B model but each layer in the 248M parameter model has 1/2 the width and each layer in the 71M parameter model 1/3 the width. The 71M model was trained with an initial learning rate of 0.001586, batch size of 480, and weight decay of 0.044506. The 248M model had an initial learning rate of 0.001831, batch size of 640, and weight decay of 0.051376.

\begin{figure}[h]
\begin{subfigure}{\textwidth}
    \centering
    \includegraphics[width=\linewidth]{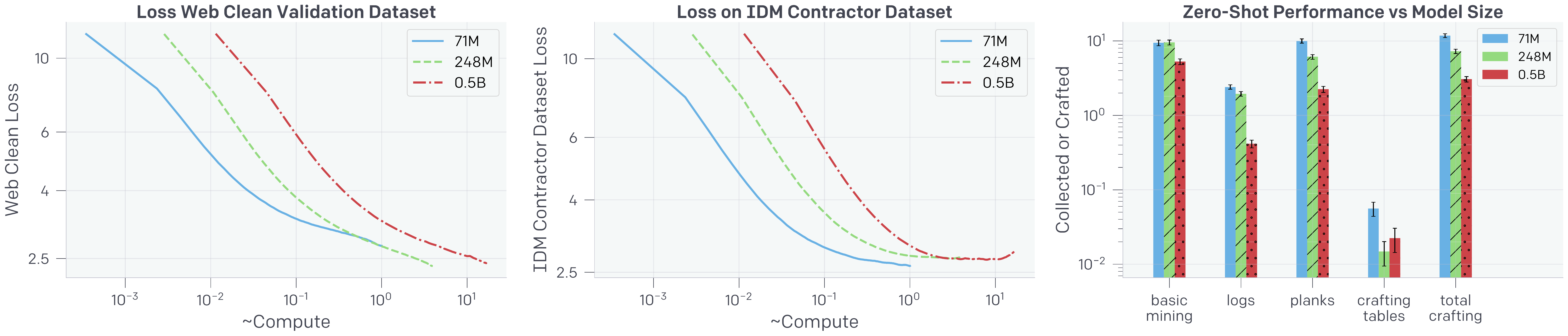} 
\end{subfigure}
\begin{subfigure}{\textwidth}
    \centering
    \includegraphics[width=\linewidth]{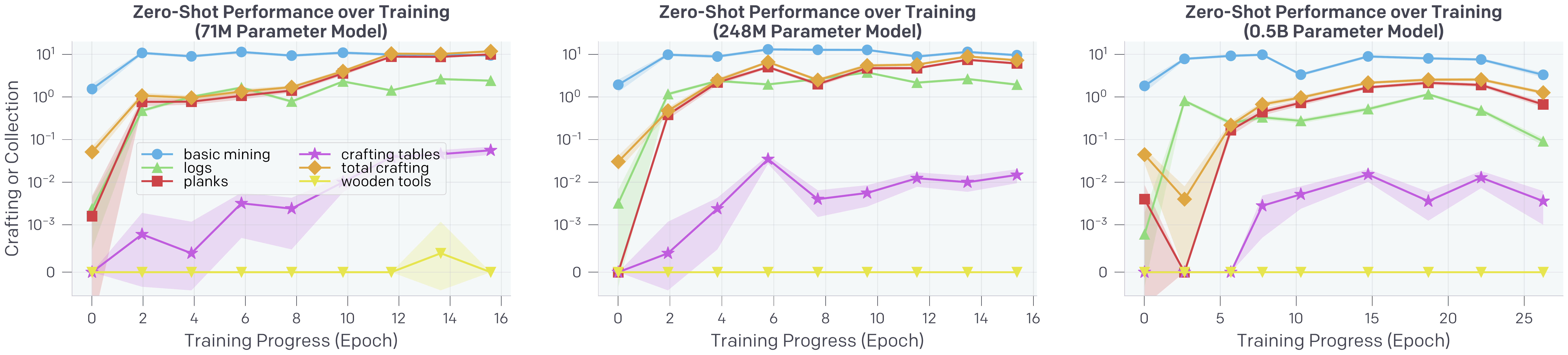} 
\end{subfigure}
\caption{
Training and Zero-Shot Performance versus Model Scale. In the first two plots the x-axis is  compute normalized to that used by the 71M parameter model, such that after 15 epochs of training the 71M model has used 1 "compute".  The 248M parameter model and the 71M model are trained on the same amount of data (15 epochs), and the 0.5B parameter model is trained on 30 epochs of data.
\textbf{(Top Left)} Loss on the \texttt{web\_clean} validation dataset.
\textbf{(Top Middle)} Loss on the IDM contractor dataset; note that these models were trained only on \texttt{web\_clean} and not on any contractor data.
\textbf{(Top Right)} Zero-shot environment rollout performance at the end of training.
\textbf{(Bottom)} Zero-shot environment rollout performance over training for the 71M model (bottom left), 248M model (bottom middle), and 0.5B model (bottom right).
}
\label{fig:model_scale_zeroshot}
\end{figure}

In Figure~\ref{fig:model_scale_zeroshot} we show validation loss on \texttt{web\_clean} with IDM pseudo-labels, loss on the contractor dataset used to train the IDM with ground truth labels collected during contractor play, and zero-shot environment performance for the 71M, 248M, and 0.5B models. While larger models have better validation loss on \texttt{web\_clean}, these results do not tell the clear story that the 0.5B model is better than its smaller counterparts. The 71M model has the lowest contractor dataset loss while having the highest \texttt{web\_clean} loss, and it also has the best zero-shot environment performance. In fact, we see that the 71M model even had non-zero wooden tool crafting (Fig.~\ref{fig:model_scale_zeroshot} bottom left). The 248M model also appears to be better at crafting than the 0.5B, and also has lower contractor dataset loss.

\begin{figure}[h]
\centering
\includegraphics[width=\linewidth]{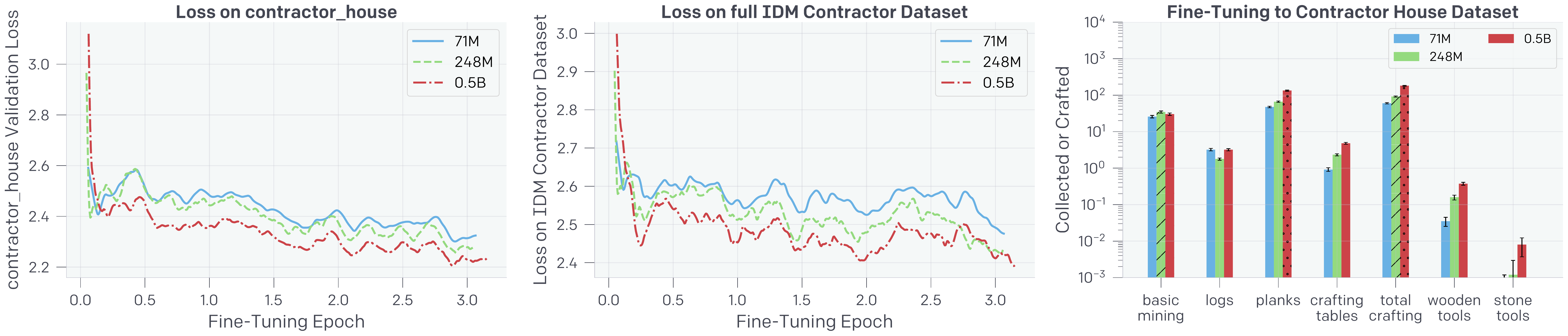} 
\caption{
\texttt{contractor\_house} fine-tuning performance versus model size.
\textbf{(Left)} Loss on the \texttt{contractor\_house} holdout validation set.
\textbf{(Middle)} Loss on the full contractor dataset collected to train the IDM; this dataset is disjoint from \texttt{contractor\_house}.
\textbf{(Right)} Environment rollout performance at the end of fine-tuning.
}
\label{fig:model_scale_ft}
\end{figure}

While the zero-shot results suggest smaller models are better, fine-tuning tells another story.
When fine-tuning to \texttt{contractor\_house}, model size rank ordering reverses and now the 0.5B model performs best both in validation loss (Fig.~\ref{fig:model_scale_ft} left) and in environment performance (Fig.~\ref{fig:model_scale_ft} right) followed by the 248M model and then the 71M model.
Environment model rollouts are performed using the same game engine that we use to collect contractor data, which could be visually distinct from videos taken from the web.
It is plausible that the larger models overfocus on the visual peculiarities in web data during pretraining since they have worse contractor data loss (Fig.\ref{fig:model_scale_zeroshot} top middle), and this causes them to perform more poorly in the environment zero-shot. However, we hypothesize that because the \texttt{contractor\_house} dataset we fine-tune to is collected from our game engine, the larger models that are a better overall Minecraft prior (as indicated by lower \texttt{web\_clean} validation loss in Fig.\ref{fig:model_scale_zeroshot} top left) can quickly shift their low level features to perform better on data coming from our game engine, resulting in better environment rollout performance. This hypothesis is further supported by Fig.~\ref{fig:model_scale_ft} (middle) showing loss on the contractor dataset collected for IDM training, which has no overlap with \texttt{contractor\_house}. After just a few steps of fine-tuning to \texttt{contractor\_house}, all models quickly improve in loss on the full IDM contractor dataset, with larger models now performing best. While not conclusive, we believe this investigation provides some intuition for future studies of model scaling for sequential decision making problems.

\section{Text Conditioning}
\label{appendix:text_conditioning}

Goal-conditioned policies~\cite{andrychowicz2017hindsight,schaul2015universal} make it possible for a single agent to perform a wide variety of goals in a single environment, which is particularly relevant in open-ended environments such as Minecraft. In recent work, goal specification has increasingly taken the form of domain specific languages~\cite{team2021open}, or even natural language~\cite{luketina2019survey,team2021creating}. The benefits of language-conditioned agents can be tremendous, especially \emph{natural}-language-conditioned agents, as their goal space contains a wide variety of potentially very complex tasks. Text conditional models have shown an amazing ability to perform tasks zero-shot (or learn them few-shot) including generalizing in impressive ways via the compositional and combinatorial possibilities allowed by natural language (e.g.\ GPT\cite{brown2020language} and DALL·E 2\cite{ramesh2022hierarchical}). We hypothesize that we should expect similar capabilities to emerge with natural-language-conditioned virtual agents, if they are similarly trained on enormous amounts of data (that goes from a natural language description to a sequence of actions that completes the specified goal). In this section we take preliminary steps toward that future. Our preliminary experiments provide evidence that it is possible to pretrain a natural-language-conditioned model for Minecraft using the general approach presented in this paper (VPT) plus conditioning on the speech that often accompanies videos.

In online videos, the human actor sometimes indicates their intent in their verbal commentary (e.g.\ ``Let's go chop some trees to make a wooden axe'' or ``now let's learn how to crop photos in Photoshop''). Conditioning on this closed caption data could produce a \emph{steerable} pre-trained model: i.e., it may later be possible to condition the model with text such as ``I am going to craft a wooden pickaxe'' or ``I am going to build a house,'' and have the agent perform those tasks specifically rather than simply follow typical human behavior (as was investigated in the rest of this paper). An alternate way to produce a steerable agent is via RL fine-tuning, which we could have done in Section \ref{section:RL_finetuning} by adding a bit indicating the task to be completed, as has been done in prior work \cite{kanitscheider2021multi}. However, conditioning on natural language offers many benefits over that approach. First, it is flexible and powerful, being able to express any task. Second, one does not need to preconceive of the task to be completed ahead of time. This would allow for general, capable, zero-shot agents like GPT, but extending those capabilities to embodied tasks such as completing tasks on computers or in simulated 3D worlds. Third, text conditioning can be used even when tasks are difficult to specify via reward functions (e.g.\ ``Let's build a house'' or--if the agent is capable of doing it--more complex things like ``I will now build a castle surrounded by a moat''). In the limit, VPT+text could conceivably produce powerful, capable, natural-language-conditional agents with the powers of GPT to meta-learn, follow instructions, and complete tasks zero or few shot, but in the form of agents that can act in virtual worlds, complete tasks on computers, and in other similar embodied sequential decision domains. We do not reach those lofty goals in this work, but we began a first step towards exploring in that direction. 

Many Minecraft videos feature audio commentary from the player. This commentary is sometimes present in the form of closed captions for the videos, or could be extracted post-hoc using automated speech recognition (ASR).\cite{yu2016automatic} Our dataset features about 17k
%\ale{FYI: more exact number = 17651. The jsonnet says 15597 for some reason} 
hours of content with associated closed captions.

We fine-tuned the 220 million parameter VPT foundation model used in the RL-fine-tuning experiments  (chosen vs.\ 0.5B for the same reason: to reduce compute costs) 
%\ale{Note: a model is named here, we should keep the name up to date with whatever we end up doing. Exact name of the base model for context: \texttt{bowen-yt-bc-2xw-allyt-20220311-1}.}
with an additional text-conditioning input on the subset of our data for which closed captions are available. To obtain the conditioning input, we first split videos into 30 second chunks. The same text is associated with every frame in a given chunk, and is made up of all the closed captions occurring within that chunk, as well as the line of text preceding and following the chunk (if any). Because the vast majority (around 95\%) of our closed caption data lacks capitalization and punctuation, it is punctuated using the rpunct library\cite{Daulet2021rpunct}. We then obtain a text embedding vector of length 4,096 from the OpenAI embedding API\cite{neelakantan2022text}, which is processed by a randomly initialized multi-layer perceptron (MLP) with two hidden layers of size 2,048. The resulting activations are added for each frame to the pretrained model activations before the transformer layers (\texttt{pretransformerActivations += mlp(textEmbedding)}).
The model is fine-tuned for four epochs.

\begin{figure}[htbp]
\begin{center}
\begin{overpic}[width=0.42\linewidth]{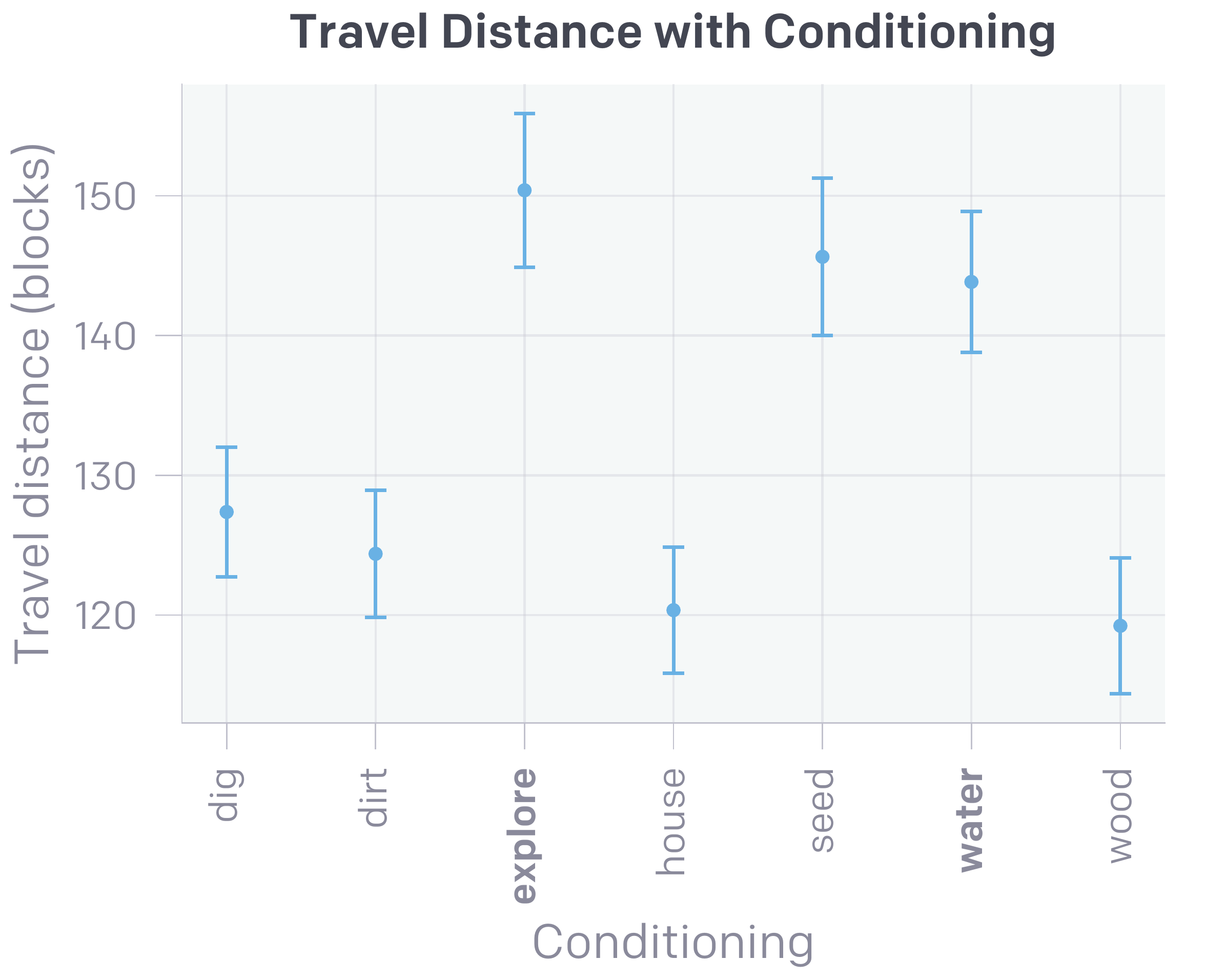}
\put(0, 78){\tiny(\textbf{a})}
\end{overpic}
\hspace{5mm}
\begin{overpic}[width=0.42\linewidth]{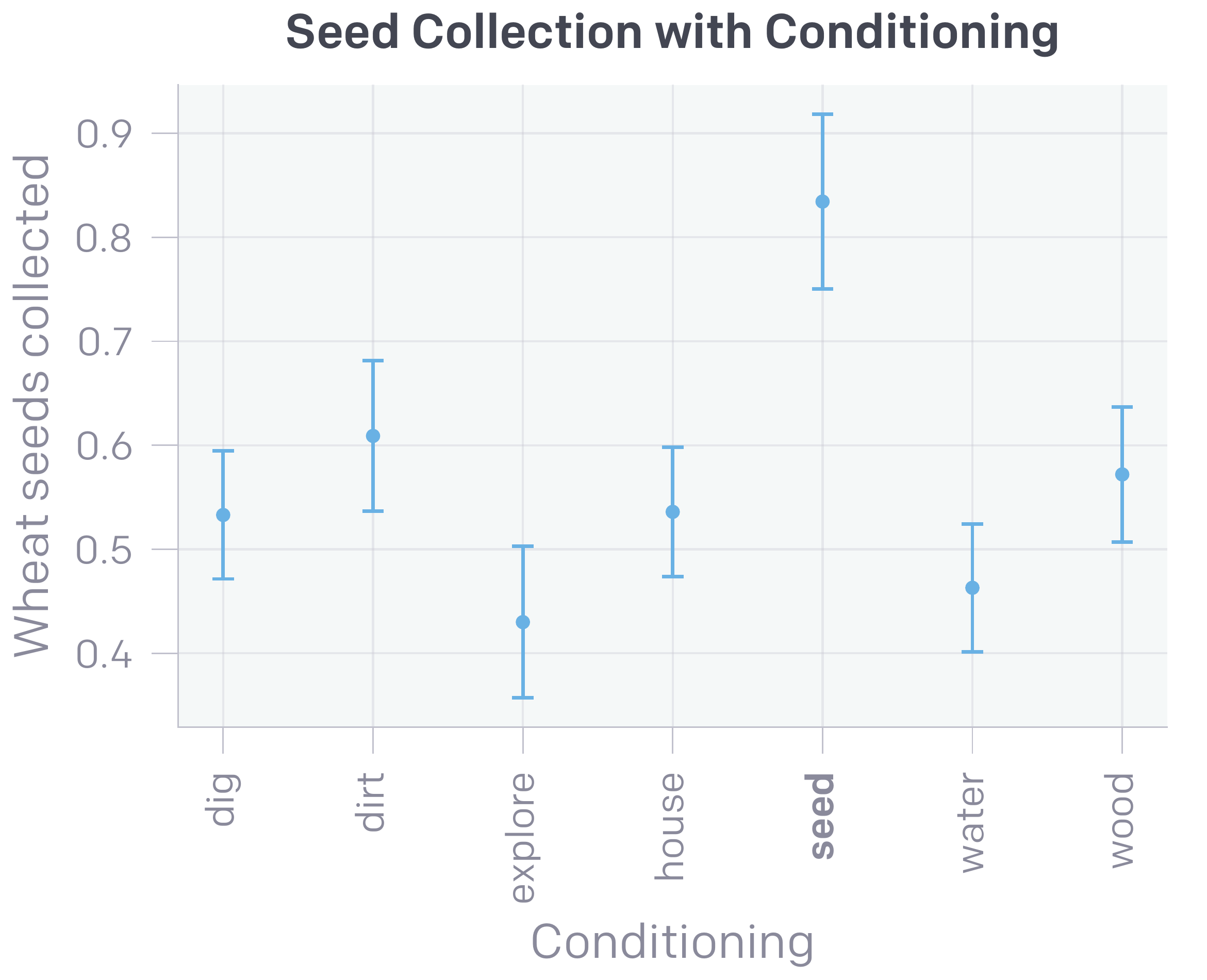}
\put(0, 78){\tiny(\textbf{b})}
\end{overpic}\\
\vspace{5mm}
\begin{overpic}[width=0.42\linewidth]{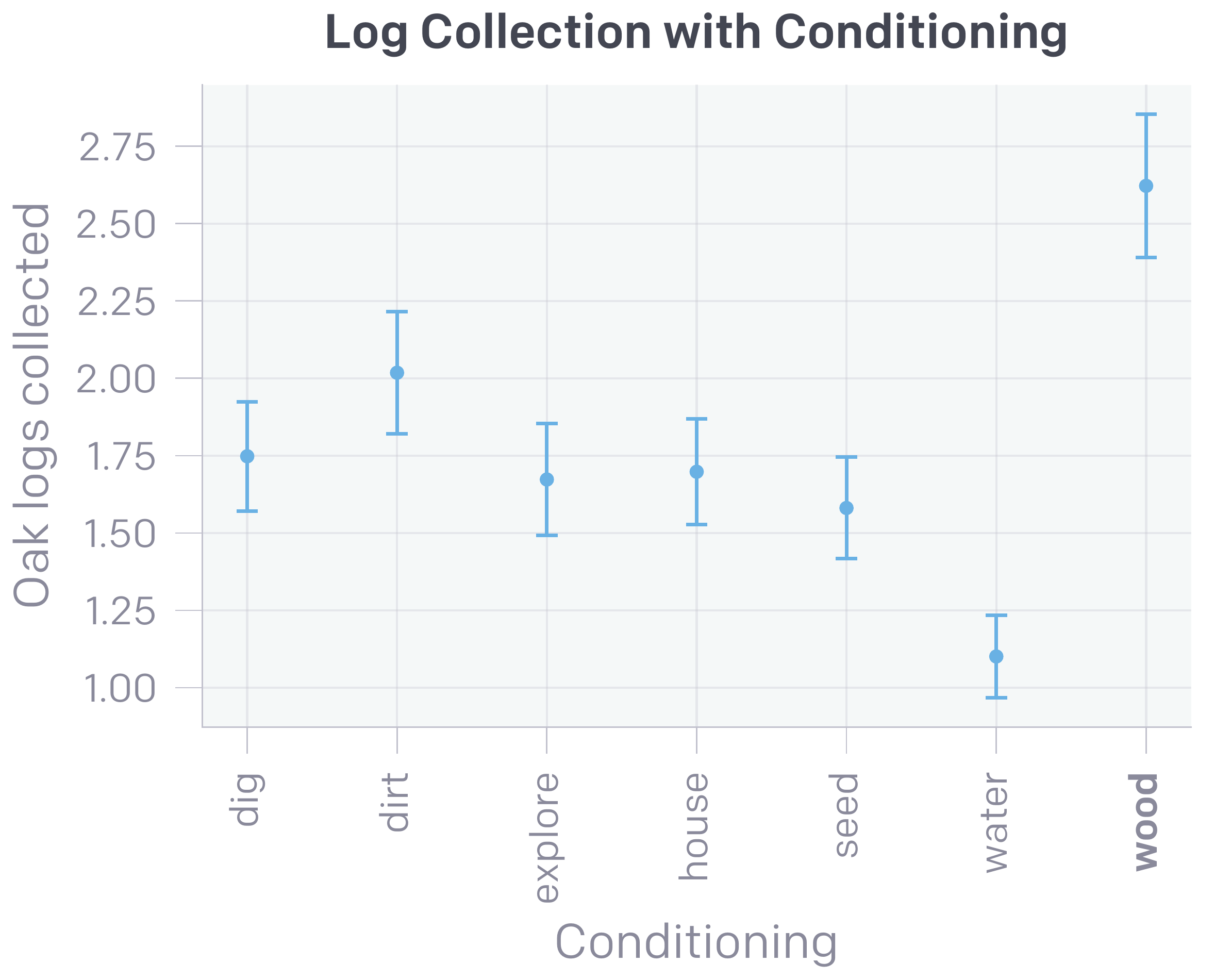}
\put(0, 78){\tiny(\textbf{c})}
\end{overpic}
\hspace{5mm}
\begin{overpic}[width=0.42\linewidth]{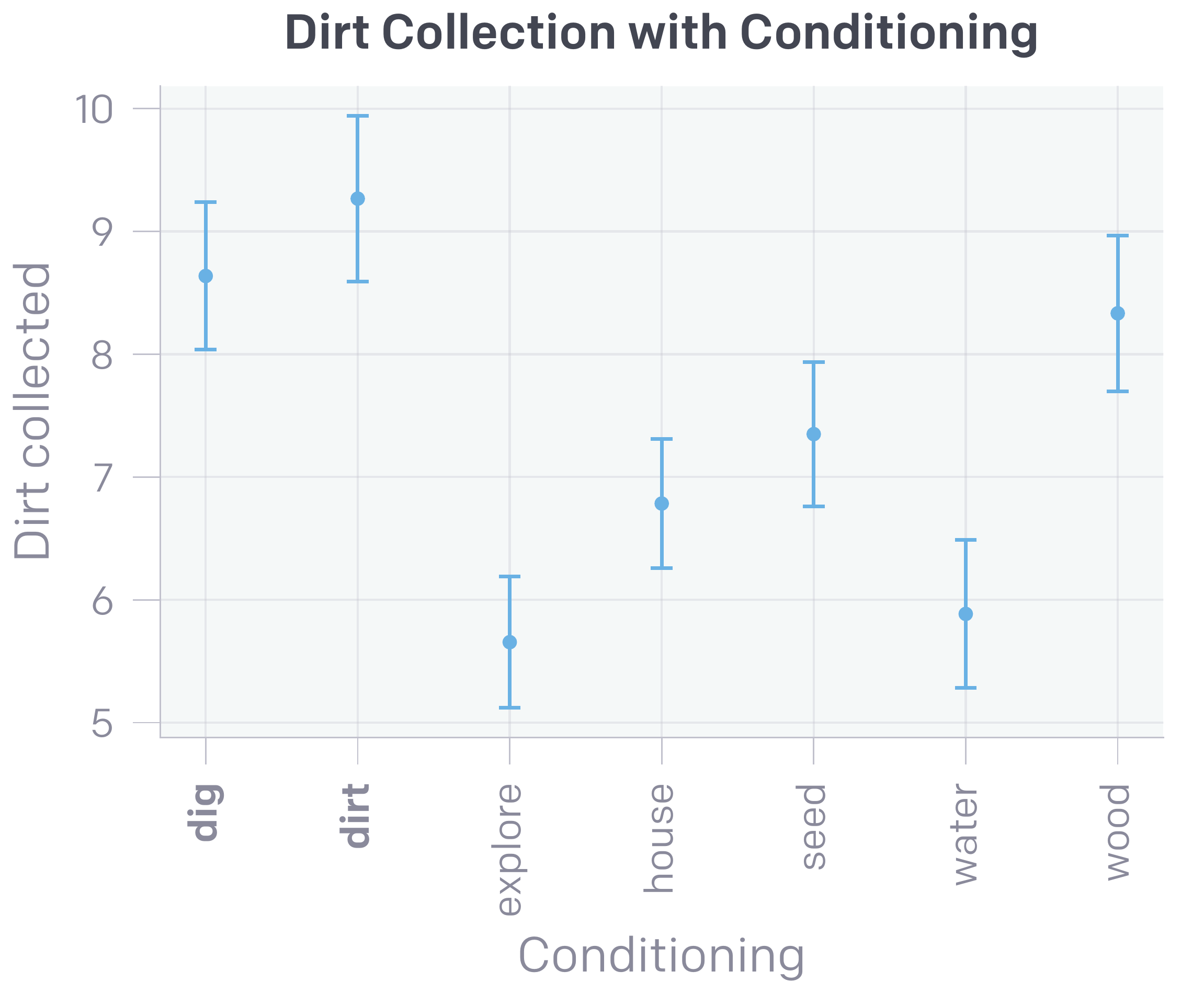}
\put(0, 78){\tiny(\textbf{d})}
\end{overpic}
\end{center}
\caption{Evidence for conditioning. In each plot, the variants expected to stand out are shown in bold. The strings corresponding to each variant are shown in Table~\ref{tab:cond_strings}. Statistics are measured over 5 minute episodes.
\textbf{(a)} Distance traveled by the agent . Both ``explore'' and ``water'' text strings should encourage a steerable agent to move more than when doing other tasks, which is what occurs. Grass (which is needed to get seeds) is not present in all biomes, which is likely why the ``seed'' condition produces more travel (as the agent sometimes needs to move to a biome with grass). The travel distance is the Euclidean distance from the spawn point to the farthest point the agent reached during the episode on the horizontal (x-z) plane. 
\textbf{(b)} Collection of wheat seeds. The ``seed'' variant collects substantially more than other variants, as expected of a steerable agent.
\textbf{(c)} Collection of oak (the most common type of wood) logs. The ``wood'' variant collects significantly more oak logs, as is to be expected of a steerable agent (we speculate that the ``water'' variant collects less because there are no trees in water). 
\textbf{(d)} Collection of dirt. The ``dirt'' and ``dig'' variants collect a large amount, and are the variants that are (indirectly in the case of ``dig'') conditioned to collect dirt. It is easy to mistakenly aim at the ground rather than at grass or trees when collecting seeds or wood, which likely explains the slightly higher amount of dirt collected by these variants. In all cases, the error bars are 95\% confidence intervals of the mean, over 1,000 episodes per conditioning variant. Treatments for which the bars in each bar plot do not overlap are statistically significantly different at a $p<0.05$ level. }
\label{fig:cond_travel_xz}
\end{figure}

\begin{table}[htbp]
\begin{center}
    \begin{tabular}{|c|c|}
        \hline
        \textbf{Variant name} & \textbf{String} \\
        \hline
        dig & I'm going to dig as far as possible \\
        dirt & I'm going to collect dirt \\
        explore & I'm going to explore \\
        house & I'm going to make a house \\
        seed & I'm going to collect seeds \\
        water & I'm going to find water \\
        wood & I'm going to chop wood \\
        \hline
    \end{tabular}
\end{center}
\caption{Strings corresponding to each conditioning variant.}
\label{tab:cond_strings}
\end{table}

Our model shows evidence of steerability. When conditioned on sentences that incite the agent to explore (such as ``I'm going to explore'' and ``I'm going to find water'') the agent travels significantly farther from its spawn point (Figure~\ref{fig:cond_travel_xz}a). Additionally, we can steer the agent to preferentially collect early game items such as seeds, wood, and dirt by conditioning with text such as ``I'm going to collect seeds/chop wood/collect dirt'' (Figure~\ref{fig:cond_travel_xz}b,c,d).

While our results show some level of steerability, more work is required to increase it. For example, we were not able to successfully steer agents to gather flowers or to hunt, both of which are possible in the early game, but less common (and, in the case of hunting animals, much more difficult) than gathering dirt, wood, or seeds. Likewise, an experiment in which the agent is presented with a crafting window and various resources, and conditioned to craft a given item (e.g.\ ``I'm going to craft a wooden axe'') failed to show that the conditioning had a significant effect on which items got crafted. Instead, it seemed the agent was more influenced by the prior, unconditional probability of what human players would craft next given the resources available, which is not too surprising since in Minecraft, especially in the early game, there is a relatively consistent path to gathering resources in a specific order go produce more powerful tools (Fig.~\ref{fig:rlft_curriculum}). For example, if the agent had the resources to make a stone pickaxe and we asked it instead to make a (weaker) wooden pickaxe, it often would make the stone pickaxe anyway. Finally, looking at videos of agent behaviors failed to convince us that the ``house'' conditioning causes the agents to take more steps towards building a house than other variants. 

Thus, our results show that it is possible to train a somewhat steerable natural-language-conditioned agent. However, its steerability is still too weak to be practically useful, and it is far from what we believe could be accomplished with more research, data, and training compute. Another exciting research direction is to have the model predict future text as well as just the next action.

\end{document}